\documentclass{article}

\PassOptionsToPackage{numbers, compress}{natbib}

\usepackage[final]{neurips_2020}

\usepackage[utf8]{inputenc} % allow utf-8 input
\usepackage[T1]{fontenc}    % use 8-bit T1 fonts
\usepackage{url}            % simple URL typesetting
\usepackage{amsfonts}       % blackboard math symbols
\usepackage{nicefrac}       % compact symbols for 1/2, etc.
\usepackage{microtype}      % microtypography

\usepackage{booktabs} % for professional tables

\usepackage{hyperref}

\hypersetup{
pdftitle={Ultrahyperbolic Representation Learning},
pdfsubject={Ultrahyperbolic Representation Learning}
}

\usepackage{mathrsfs}  
\usepackage[table]{xcolor}
\usepackage{xcolor}
\usepackage{dsfont}
\usepackage{xspace}
\usepackage{listings}
\usepackage{numprint}
\usepackage{multirow}
\usepackage{setspace,lipsum}
\usepackage{graphicx}
\usepackage{subfigure}

\usepackage{amsmath, amsthm}
\usepackage{amssymb}

\usepackage{centernot}
\usepackage{algorithm}
\usepackage{algpseudocode}
\newcommand{\ie}{\textit{i.e.}\xspace}
\newcommand{\eg}{\textit{e.g.}\xspace}
\newcommand{\wrt}{\textit{w.r.t.}\xspace}

\newcommand{\avec}{\mathbf{a}}
\newcommand{\bvec}{\mathbf{b}}
\newcommand{\cvec}{\mathbf{c}}
\newcommand{\dvec}{\mathbf{d}}

\newcommand{\uvec}{\mathbf{u}}
\newcommand{\vvec}{\mathbf{v}}

\newcommand{\svec}{\mathbf{s}}
\newcommand{\tvec}{\mathbf{t}}
\newcommand{\zvec}{\mathbf{z}}
\newcommand{\xvec}{\mathbf{x}}
\newcommand{\yvec}{\mathbf{y}}

\newcommand{\Dcal}{\mathcal{D}}
\newcommand{\Hcal}{\mathcal{H}}
\newcommand{\Qcal}{\mathcal{Q}}

\newcommand{\Mcal}{\mathcal{M}}
\newcommand{\Ncal}{\mathcal{N}}
\newcommand{\Scal}{\mathcal{S}}
\newcommand{\Ucal}{\mathcal{U}}
\newcommand{\Wcal}{\mathcal{W}}
\newcommand{\RR}{\mathbb{R}}
\newcommand{\RRR}{\mathbb{R}^{d}}
\newcommand{\RRpseudo}{\mathbb{R}^{p,q+1}}
\newcommand{\dsf}{\mathsf{d}}
\newcommand{\Dsf}{\mathsf{D}}

\newcommand{\xii}{\boldsymbol{\xi}}
\newcommand{\chii}{\boldsymbol{\chi}}
\newcommand{\zetaa}{\boldsymbol{\zeta}}

\newcommand{\hquadric}{\Qcal^{p,q}_{\beta}}
\newcommand{\hquadricpq}[3]{\Qcal^{#1,#2}_{#3}}
\newcommand{\sqnorm}[1]{\| #1 \|_q^2}

\newcommand{\tangentspace}[2]{T_{#1} #2}
\newcommand{\tquadric}[1]{\tangentspace{#1} \hquadric}
\newcommand{\ambientdistance}[2]{\dsf_q (#1, #2)}
\newcommand{\geodistance}[2]{\dsf_\gamma (#1, #2)}
\newcommand{\ourdistance}[2]{\Dsf_\gamma (#1, #2)}

\newcommand{\preconditioner}{\textbf{P}_{\xvec}}

\newcommand{\sridentitymatrix}{\textbf{I}_{q+1,p}}
\newcommand{\srbasis}{\textbf{G}}
\newcommand{\zeroo}{\textbf{0}}
\newcommand{\differential}{\textrm{d}f}
\newcommand{\covariantderivative}{\textrm{D}}

\newcommand{\orthoproji}[1]{\Pi_{#1}}
\newcommand{\orthoproj}{\orthoproji{\xvec}}
\newcommand{\graph}{G}
\newcommand{\nodes}{V}
\newcommand{\edges}{E}
\newcommand{\Capacity}{\textbf{C}}
\newcommand{\SymCapacity}{\textbf{S}}

\newcommand{\Xvec}{\xii}
\newcommand{\Yvec}{\zetaa}

\newcommand{\inner}[2]{\langle #1 , #2 \rangle}
\newcommand{\innerlorentzq}[3]{\langle #1 , #2 \rangle_{#3}}
\newcommand{\dinnerlorentz}[1]{\langle #1 , #1 \rangle_q}

\newcommand{\innerlorentz}[2]{\innerlorentzq{#1}{#2}{q}}
\newcommand{\eucg}{\nabla f (\xvec)}
\newcommand{\eucgi}{\nabla f (\xvec_i)}
\newcommand{\descent}{\chii}
\newcommand{\srgrad}{D f (\xvec)}
\newcommand{\Poincare}{Poincar\'e~}

\newcommand{\overbar}[1]{\mkern 1.0mu\overline{\mkern-0.7mu#1\mkern-1.0mu}\mkern 1.0mu}

\newcommand{\MMcal}{\overbar{\Mcal}}
\newtheorem{theorem}{Theorem}[section]

\title{Ultrahyperbolic Representation Learning}

\author{%
  Marc T. Law ~~~~~~~~~~~~~~~~~~~~~~~~~~~~~~~~ Jos Stam \\ 
  ~ \\
  NVIDIA 
}

\begin{document}

\maketitle

\begin{abstract}

In machine learning, data is usually represented in a (flat) Euclidean space where distances between points are along straight lines. Researchers have recently considered more exotic (non-Euclidean) Riemannian manifolds such as hyperbolic space which is well suited for tree-like data. In this paper, we propose a representation living on a pseudo-Riemannian manifold of constant nonzero curvature. It is a generalization of hyperbolic and spherical geometries where the nondegenerate metric tensor need not be positive definite. We provide the necessary learning tools in this geometry and extend gradient-based optimization techniques. More specifically, we provide closed-form expressions for distances via geodesics and define a descent direction to minimize some objective function. Our novel framework is applied to graph representations. 

\end{abstract}

\section{Introduction} \label{sec:introduction}

In most machine learning applications, data representations lie on a smooth manifold \cite{lee2013introduction} and the training procedure is optimized with an iterative algorithm such as line search or trust region methods~\cite{nocedal2006numerical}. 
In most cases, the smooth manifold is Riemannian, which means that it is equipped with a positive definite metric. Due to the positive definiteness of the metric, the negative of the (Riemannian) gradient is a descent direction that can be exploited to iteratively minimize some objective function \cite{absil2009optimization}. 

The choice of metric on the Riemannian manifold determines how relations between points are quantified. The most common Riemannian manifold is the flat Euclidean space, which has constant zero curvature and the distances between points are measured by straight lines. An intuitive example of non-Euclidean Riemannian manifold is the spherical model (\ie representations lie on a sphere) that has constant positive curvature and is used for instance in face recognition \cite{Tapaswi_2019_ICCV,wang2017normface}. On the sphere, geodesic distances are a function of angles. Similarly, Riemannian spaces of constant negative curvature are called hyperbolic \cite{petersen2006riemannian}. Such spaces were shown by Gromov to be well suited to represent tree-like structures \cite{gromov1987hyperbolic}. The machine learning community has adopted these spaces to learn tree-like graphs \cite{chen2013hyperbolicity} and hierarchical data structures \cite{pmlr-v97-law19a,nickel2017poincare,nickel2018learning}, and also to compute means in tree-like shapes \cite{feragen2011means,feragen2012toward}.

In this paper, we consider a class of pseudo-Riemannian manifolds of constant nonzero curvature \cite{wolf1972spaces} not previously considered in machine learning. These manifolds not only generalize the hyperbolic and spherical geometries mentioned above, but also contain hyperbolic and spherical submanifolds and can therefore describe relationships specific to those geometries. 
The difference is that we consider the larger class of pseudo-Riemannian manifolds where the considered nondegenerate metric tensor need not be positive definite. Optimizing a cost function on our non-flat ultrahyperbolic space requires a descent direction method that follows a path along the curved manifold. We achieve this by employing tools from differential geometry such as geodesics and exponential maps. 
The theoretical contributions in this paper are two-fold: (1) explicit methods to calculate dissimilarities and (2) general optimization tools on pseudo-Riemannian manifolds of constant nonzero curvature.

\begin{figure}[!t]
    \centering
    \includegraphics[width=.35\linewidth]{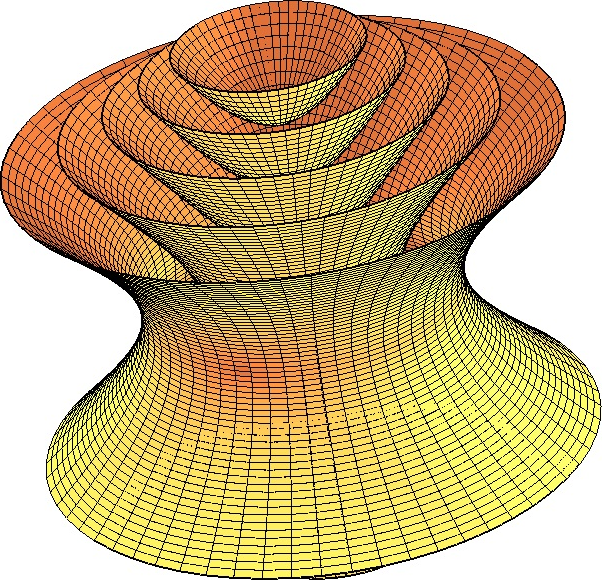} ~~~~~~~~~~~~~~
    \includegraphics[width=.35\linewidth]{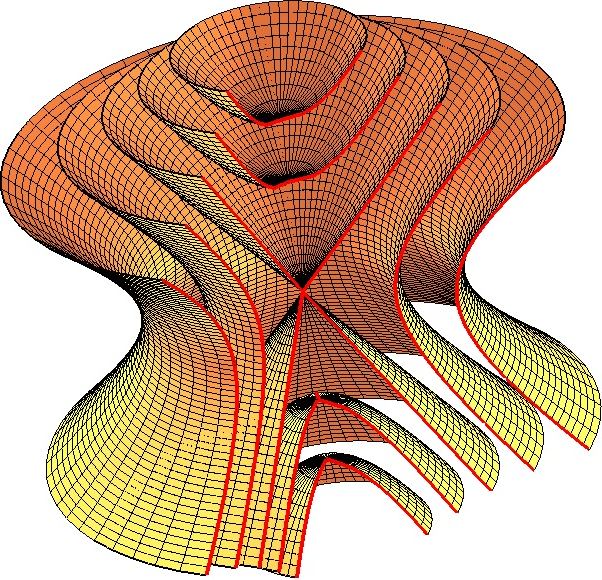} 
    \caption{Two iso-surface depictions of an orthographic projection of the same pseudo-hyperboloid $\Qcal^{2,1}_{-1}$ into $\RR^3$ along one time dimension. It contains the entire family of hyperboloids as submanifolds.} 
    \label{fig:ultrahyperbolic}
\end{figure}

\section{Pseudo-Hyperboloids} \label{sec:background}

\textbf{Notation:} We denote points on a smooth manifold $\Mcal$ \cite{lee2013introduction} by boldface Roman characters $\xvec \in \Mcal$. $T_{\xvec} \Mcal$ is the tangent space of $\Mcal$ at $\xvec$ and we write tangent vectors $\Xvec \in T_{\xvec} \Mcal$ in boldface Greek fonts. 
$\RR^d$ is the (flat) $d$-dimensional Euclidean space, it is equipped with the (positive definite) dot product denoted by $\langle \cdot , \cdot \rangle$ and defined as $\langle \xvec , \yvec \rangle = \xvec^{\top}\yvec$. 
The $\ell_2$-norm of $\xvec$ is $\| \xvec \|=\sqrt{\langle \xvec , \xvec \rangle}$.  $\RR^d_* = \RR^d \backslash \{ \boldsymbol{0} \}$ is the Euclidean space with the origin removed.

\textbf{Pseudo-Riemannian manifolds:} A smooth manifold $\Mcal$ is pseudo-Riemannian (also called semi-Riemannian \cite{o1983semi}) if it is equipped with a pseudo-Riemannian metric tensor (named ``metric'' for short in differential geometry). The pseudo-Riemannian metric $g_{\xvec} : T_{\xvec} \Mcal \times T_{\xvec} \Mcal \to \RR$ at some point $\xvec \in \Mcal$ is a nondegenerate symmetric bilinear form. Nondegeneracy means that if for a given $\Xvec \in T_{\xvec} \Mcal$ and for all $\Yvec \in T_{\xvec} \Mcal$ we have $g_{\xvec}(\Xvec, \Yvec) = 0$, then $\Xvec = \zeroo$. 
If the metric is also positive definite (\ie $\forall \Xvec \in T_{\xvec} \Mcal,~ g_{\xvec}(\Xvec, \Xvec) > 0$ iff $\Xvec \neq \zeroo$), then it is Riemannian. Riemannian geometry is a special case of pseudo-Riemannian geometry where the metric is positive definite. In general, this is not the case and non-Riemannian manifolds distinguish themselves by having some non-vanishing tangent vectors $\Xvec \neq \zeroo$ that satisfy $g_{\xvec}(\Xvec,\Xvec) \leq 0$. 
We refer the reader to \cite{anciaux2011minimal,o1983semi,wolf1972spaces} for details.

\textbf{Pseudo-hyperboloids} generalize spherical and hyperbolic manifolds to the class of pseudo-Riemannian manifolds. Let us note $d=p+q+1 \in \mathbb{N}$ the dimensionality of some pseudo-Euclidean space where each vector is written $\xvec=\left(x_0, x_1, \cdots, x_{q+p}\right)^{\top}$. 
That space is denoted by $\RRpseudo$ when it is equipped with the following scalar product (\ie nondegenerate symmetric bilinear form \cite{o1983semi}):
\begin{equation} \label{eq:q_lorentz_inner_product}
\forall \avec = \left(a_0, \cdots, a_{q+p}\right)^{\top}, ~ \bvec = \left(b_0, \cdots, b_{q+p}\right)^{\top}, ~ \innerlorentz{\avec}{\bvec} = - \sum_{i=0}^q a_i b_i + \sum_{j=q+1}^{p+q} a_j b_j = \avec^{\top} \srbasis \bvec,
\end{equation}
where $\srbasis = \srbasis^{-1} = \sridentitymatrix$ is the $d\times d$ diagonal matrix with the first $q+1$ diagonal elements equal to $-1$ and the remaining $p$ equal to $1$. Since $\RRpseudo$ is a vector space, we can identify the tangent space to the space itself by means of the natural isomorphism $T_{\xvec} \RRpseudo \approx \RRpseudo$. Using the terminology of special relativity, $\RRpseudo$ has $q+1$ time dimensions and $p$ space dimensions.

A pseudo-hyperboloid is the following submanifold of codimension one (\ie hypersurface) in $\RRpseudo$:
\begin{equation}
    \Qcal^{p,q}_\beta = \left\{ \xvec = \left(x_0, x_1, \cdots, x_{p+q}\right)^{\top} \in \RRpseudo: \sqnorm{\xvec} = \beta \right\},
\end{equation}
where $\beta \in \RR_*$ is a nonzero real number and the function $\sqnorm{\cdot}$ given by $\sqnorm{\xvec} = \langle \xvec , \xvec \rangle_{q}$ is the associated quadratic form of the scalar product. 
It is equivalent to work with either $\Qcal^{p,q}_\beta$ or $\Qcal^{q+1,p-1}_{-\beta}$ as they are interchangeable via an anti-isometry  (see supp. material). 
For instance, the unit $q$-sphere $\Scal^{q} = \left\{ \xvec \in \RR^{q+1} : \| \xvec \| = 1 \right\}$ is anti-isometric to $\Qcal^{0,q}_{-1}$ which is then spherical. 
In the literature, the set $\hquadric$ is called a ``pseudo-sphere'' when $\beta>0$ and a ``pseudo-hyperboloid'' when $\beta<0$. In the rest of the paper, we only consider the pseudo-hyperbolic case (\ie $\beta<0$). 
Moreover, for any $\beta < 0$, $\hquadric$ is homothetic to $\hquadricpq{p}{q}{-1}$, the value of $\beta$ can then be considered to be $-1$. 
We can obtain the spherical and hyperbolic geometries by constraining all the elements of the space dimensions of a pseudo-hyperboloid to be zero or constraining all the elements of the time dimensions except one to be zero, respectively. Pseudo-hyperboloids then generalize spheres and hyperboloids.

The pseudo-hyperboloids that we consider in this paper are hard to visualize as they live in ambient spaces with dimension higher than 3. In Fig. \ref{fig:ultrahyperbolic}, we show iso-surfaces of a projection of the 3-dimensional pseudo-hyperboloid $\Qcal^{2,1}_{-1}$ (embedded in $\RR^{2,2}$) into $\RR^3$ along its first time dimension.

\textbf{Metric tensor and tangent space:} 
The metric tensor at $\xvec \in \hquadric$ is $g_{\xvec} ( \cdot, \cdot ) = \innerlorentz{\cdot}{ \cdot}$ where $g_{\xvec} : \tquadric{\xvec} \times \tquadric{\xvec} \to \RR$. By using the isomorphism $T_{\xvec} \RRpseudo \approx \RRpseudo$ mentioned above, the tangent space of $\hquadric$ at  $\xvec$ can be defined as $\tquadric{\xvec} = \left\{ \xii \in \RRpseudo : \innerlorentz{\xvec}{\xii} = 0\right\}$ for all $\beta \neq 0$. 
Finally, the orthogonal projection of an arbitrary $d$-dimensional vector $\zvec$ onto $\tquadric{\xvec}$ is:
\begin{equation} \label{eq:ortho_projection}
    \orthoproj (\zvec) = \zvec - \frac{\langle \zvec, \xvec \rangle_q}{\langle \xvec , \xvec \rangle_q} \xvec. % = 
\end{equation}

\section{Measuring Dissimilarity on Pseudo-Hyperboloids} \label{sec:pseudosphere}

This section introduces the differential geometry tools necessary to quantify dissimilarities/distances between points on $\hquadric$. 
Measuring dissimilarity is an important task in machine learning and has many applications (\eg in metric learning \cite{xing2003distance}). 

\textbf{Intrinsic geometry:} The \textit{intrinsic geometry} of the hypersurface $\hquadric$ embedded in $\RRpseudo$ (\ie the geometry perceived by the inhabitants of $\hquadric$ \cite{o1983semi}) derives solely from its metric tensor applied to tangent vectors to $\hquadric$. 
For instance, it can be used to measure the arc length of a tangent vector joining two points along a geodesic and define their geodesic distance. 
Before considering geodesic distances, we consider extrinsic distances (\ie distances in the ambient space $\RRpseudo$).
Since $\RRpseudo$ is isomorphic to its tangent space, tangent vectors to $\RRpseudo$ are naturally identified with points. Using the quadratic form of Eq.~\eqref{eq:q_lorentz_inner_product}, the extrinsic distance between two points $\avec, \bvec\in \hquadric$ is:
\begin{equation} \label{eq:ultra_hyperbolic_distance}
    \ambientdistance{\avec}{\bvec} = \sqrt{ | \sqnorm{\avec - \bvec} | } = \sqrt{ |\sqnorm{\avec} + \sqnorm{\bvec} - 2 \innerlorentz{\avec}{\bvec} | } = \sqrt{ | 2 \beta - 2 \innerlorentz{\avec}{\bvec} | }.
\end{equation}
This distance is a good proxy for the geodesic distance $\geodistance{\cdot}{\cdot}$, that we introduce below, if it preserves \textit{distance relations}: $\geodistance{\avec}{\bvec} < \geodistance{\cvec}{\dvec}$ iff $ \ambientdistance{\avec}{\bvec}  <   \ambientdistance{\cvec}{\dvec} $. This relation is satisfied for two special cases of pseudo-hyperboloids for which the geodesic distance is well known:

 $\bullet$ \textbf{Spherical manifold  ($\hquadricpq{0}{q}{\beta}$):} If $p=0$, the geodesic distance $\geodistance{\avec}{\bvec} = \sqrt{| \beta |} \cos^{-1}\left(\frac{ \innerlorentz{\avec}{\bvec}}{\beta}\right)$ is called spherical distance. 
In practice, the cosine similarity $\frac{\innerlorentz{\cdot}{\cdot}}{\beta}$ is often considered instead of $\geodistance{\cdot}{\cdot}$ since it satisfies $\geodistance{\avec}{\bvec} < \geodistance{\cvec}{\dvec}$ iff $ \innerlorentz{\avec}{\bvec}  <   \innerlorentz{\cvec}{\dvec} $ iff $ \ambientdistance{\avec}{\bvec}  <   \ambientdistance{\cvec}{\dvec}$.

 $\bullet$ \textbf{Hyperbolic manifold (upper sheet of the two-sheet hyperboloid $\hquadricpq{p}{0}{\beta}$):} If $q=0$, the geodesic distance $\geodistance{\avec}{\bvec} = \sqrt{| \beta |} \cosh^{-1}\left(\frac{\innerlorentz{\avec}{\bvec}}{\beta}\right)$ with $a_0 > 0$ and $b_0 > 0$ is called \Poincare distance \cite{nickel2018learning}.
The (extrinsic) Lorentzian distance was shown to be a good proxy in hyperbolic geometry \cite{pmlr-v97-law19a}.

For the {\bf ultrahyperbolic} case (\ie $q \geq 1$ and $p \geq 2$), the distance relations are not preserved: $\geodistance{\avec}{\bvec} < \geodistance{\cvec}{\dvec} \centernot\iff   \ambientdistance{\avec}{\bvec} <  \ambientdistance{\cvec}{\dvec}$. 
We then need to consider only geodesic distances. 
This section introduces %the first theoretical contribution of this paper, specifically 
closed-form expressions for geodesic distances on ultrahyperbolic manifolds.

\textbf{Geodesics:}
Informally, a geodesic is a curve joining points on a manifold $\Mcal$ that minimizes some ``effort'' depending on the metric. More precisely, let $I\subseteq\RR$ be a (maximal) interval containing $0$. A geodesic $\gamma:I\rightarrow\Mcal$ maps a real value $t \in I$ to a point on the manifold $\Mcal$. It is a curve on $\Mcal$ defined by its initial point $\gamma(0) = \xvec \in \Mcal$ and initial tangent vector $\gamma^\prime(0) = \Xvec \in T_{\xvec} \Mcal$ where $\gamma^\prime(t)$ is the derivative of $\gamma$ at $t$. By analogy with physics, $t$ is considered as a time value. Intuitively, one can think of the curve as the trajectory over time of a ball being pushed from a point $\xvec$ at $t = 0$ with initial velocity $\Xvec$ and constrained to roll on the manifold. We denote this curve explicitly by $\gamma_{\xvec \to \xii} (t)$ unless the dependence is obvious from the context. For this curve to be a geodesic, its acceleration has to be zero: $\forall t\in I, \gamma^{\prime\prime}(t) = \zeroo$. This condition is a second-order ordinary differential equation that has a unique solution for a given set of initial conditions \cite{lindelof1894application}. The interval $I$ is said to be maximal if it cannot be extended to a larger interval. 
In the case of $\hquadric$, we have $I = \RR$ and $I$ is then maximal.

\begin{figure}[!t]
    \centering
    \includegraphics[width=.49\linewidth]{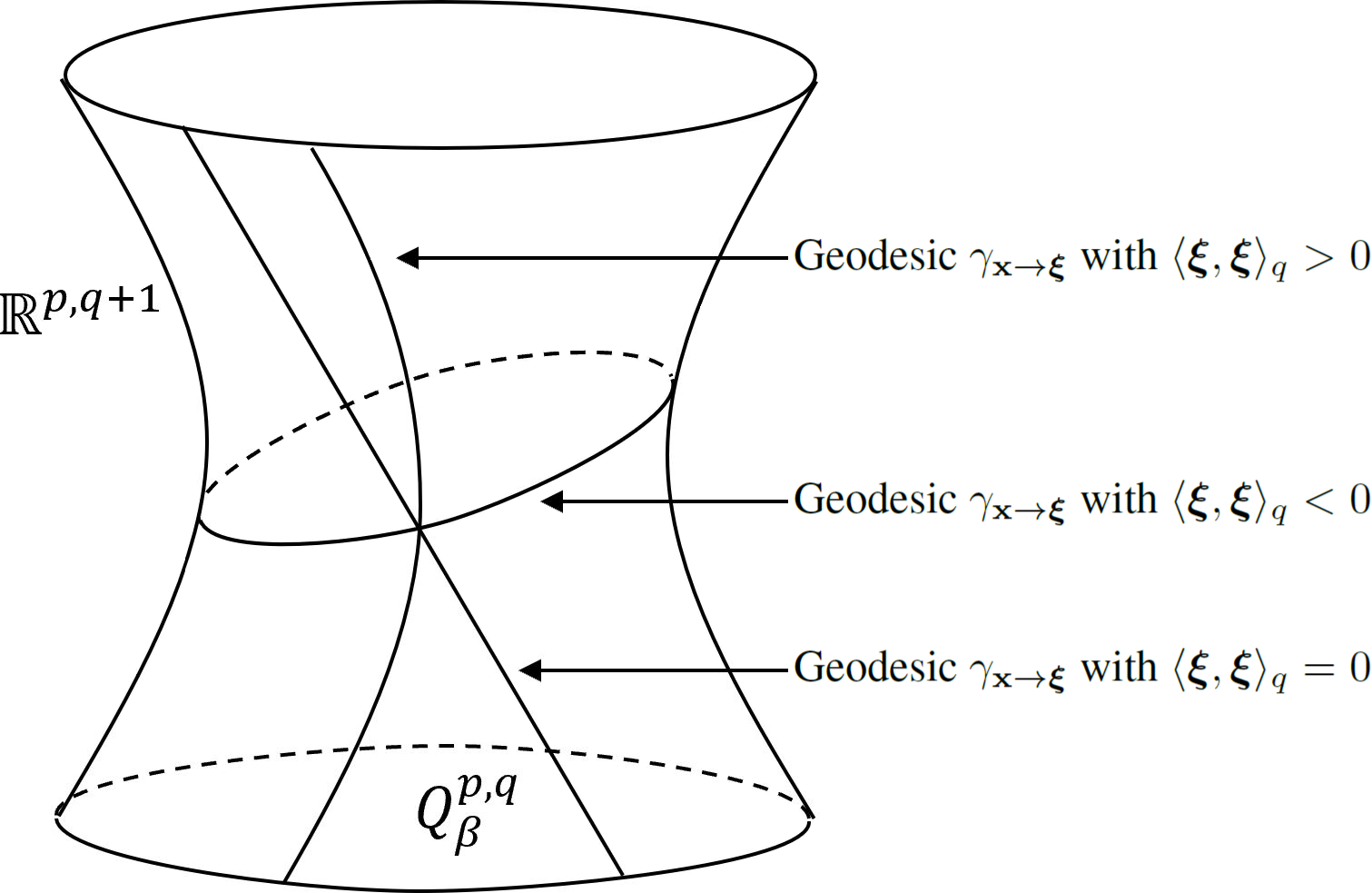} \includegraphics[width=.49\linewidth]{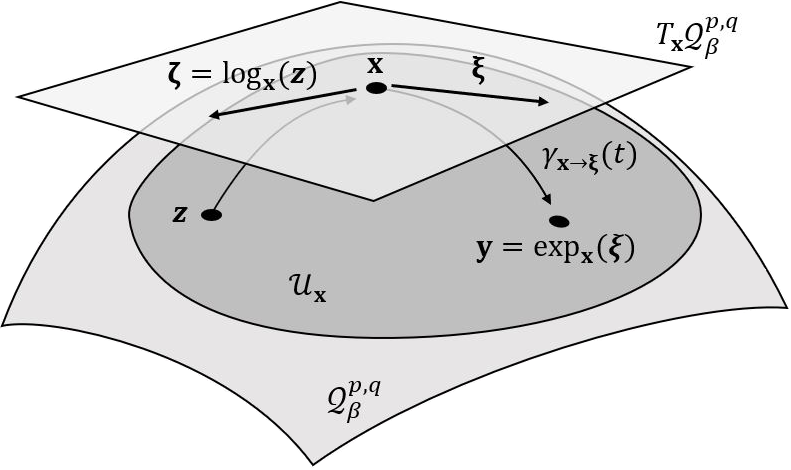} %\vspace{-1em}
    \caption{(left)  Illustration of the three types of geodesics of a pseudo-hyperboloid defined in Eq.~\eqref{eq:geodesic}. (right) Exponential map and logarithm map.}
    \label{fig:geodesics}
\end{figure}

\textbf{Geodesic of $\hquadric$:}  As we show in the supp. material, the geodesics of $\hquadric$ are a combination of the hyperbolic, flat and spherical cases. The nature of the geodesic $\gamma_{\xvec \to \xii}$ depends on the sign of $\innerlorentz{\xii}{\xii}$. For all $t \in \RR$, the geodesic $\gamma_{\xvec \to \xii}$ of $\hquadric$ with $\beta < 0$ is written:
\begin{equation} \label{eq:geodesic}
   \gamma_{\xvec \to \xii} (t) = \left\{ 
\begin{array}{l l}
 \cosh \left( \frac{t \sqrt{| \langle \xii, \xii \rangle_q |}}{\sqrt{| \beta |}} \right) \xvec +  \frac{\sqrt{| \beta |}}{\sqrt{| \langle \xii, \xii \rangle_q |}} \sinh \left(\frac{t \sqrt{| \langle \xii, \xii \rangle_q |}}{\sqrt{| \beta |}} \right) \xii & \quad \text{if } \langle \xii, \xii \rangle_q > 0\\
  \xvec + t \xii & \quad \text{if } \langle \xii, \xii \rangle_q = 0\\
  \cos \left(\frac{t \sqrt{| \langle \xii, \xii \rangle_q |}}{\sqrt{| \beta |}} \right) \xvec  ~~ + \frac{\sqrt{| \beta |}}{\sqrt{| \langle \xii, \xii \rangle_q |}} \sin \left( \frac{t \sqrt{| \langle \xii, \xii \rangle_q |}}{\sqrt{| \beta |}} \right) \xii & \quad \text{if } \langle \xii, \xii \rangle_q < 0 \end{array} \right.
\end{equation}
We recall that $\langle \xii, \xii \rangle_q = 0$ does not imply $\xii = \boldsymbol{0}$. The geodesics are an essential ingredient to define a mapping known as the exponential map. 
See Fig. \ref{fig:geodesics} (left) for a depiction of these three types of geodesics, and Fig. \ref{fig:geodesics} (right) for a depiction of the other quantities introduced in this section.

\textbf{Exponential map:}
Exponential maps are a way of collecting all of the geodesics of a pseudo-Riemannian manifold $\Mcal$ into a unique differentiable mapping.
Let $\Dcal_{\xvec}  \subseteq T_{\xvec} \Mcal$ be the set of tangent vectors $\xii$ such that $\gamma_{\xvec \to \xii}$ is defined at least on the interval $[0,1]$. This allows us to uniquely define the exponential map $\exp_{\xvec} : \Dcal_{\xvec} \to \Mcal$ such that $\exp_{\xvec} (\Xvec) = \gamma_{\xvec \to \xii} (1)$.

The manifold $\hquadric$ is geodesically complete, the domain of its exponential map is then $\Dcal_{\xvec}  = \tquadric{\xvec}$. Using Eq. \eqref{eq:geodesic} with $t=1$, we obtain an exponential map of the entire tangent space to the manifold:
\begin{equation}
\label{eq:exponential}
\forall \xii \in \tquadric{\xvec}, ~ \exp_{\xvec} (\xii) =  \gamma_{\xvec \to \xii} (1).
\end{equation}
We make the important observation that the image of the exponential map does not necessarily cover the entire manifold: not all points on a manifold are connected by a geodesic. This is the case for our pseudo-hyperboloids.  Namely, for a given point $\xvec\in\hquadric$ there exist points $\yvec$ that are not in the image of the exponential map (\ie there does not exist a tangent vector $\Xvec$ such that $\yvec = \exp_{\xvec} (\xii)$). 

\textbf{Logarithm map:} 
We provide a closed-form expression of the logarithm map for pseudo-hyperboloids.
Let $\Ucal_\xvec \subseteq \hquadric$ be some neighborhood of $\xvec$. The logarithm map $\log_\xvec  : \Ucal_\xvec \to T_\xvec \Qcal^{p,q}_{\beta}$ is defined as the inverse of the exponential map on $\Ucal_\xvec$ (\ie $\log_\xvec = \exp_\xvec^{-1}$). We propose:
\begin{equation} 
\label{eq:lambda_factor}
    \forall \yvec \in \Ucal_\xvec, ~ \log_\xvec (\yvec) = \left\{ 
\begin{array}{l l}
 \frac{\cosh^{-1}( \frac{ \langle \xvec , \yvec \rangle_q }{ \beta })}{\sqrt{ (\frac{\langle \xvec , \yvec \rangle_q}{\beta})^2 - 1}} \left( \yvec - \frac{ \langle \xvec, \yvec \rangle_q}{ \beta} \xvec \right) & \quad \text{ if }   \frac{ \langle \xvec, \yvec \rangle_q}{ | \beta |} < -1 \\ 
 \yvec - \xvec & \quad \text{ if }  \frac{ \langle \xvec, \yvec \rangle_q }{ | \beta |} = -1 \\ 
 \frac{\cos^{-1}(\frac{\langle \xvec , \yvec \rangle_q}{ \beta })}{\sqrt{ 1 - (\frac{\langle \xvec , \yvec \rangle_q}{\beta})^2}} \left( \yvec - \frac{ \langle \xvec, \yvec \rangle_q}{ \beta} \xvec \right) & \quad \text{ if } \frac{ \langle \xvec, \yvec \rangle_q }{ | \beta |} \in (-1,1)
 \end{array} \right.
\end{equation}
By substituting $\Xvec=\log_\xvec(\yvec)$ into Eq. \eqref{eq:exponential}, one can verify that our formulas are the inverse of the exponential map. 
The set $\Ucal_\xvec = \left\{ \yvec \in \hquadric : \innerlorentz{\xvec}{\yvec} < |\beta| \right\}$ is called a normal neighborhood of $\xvec \in  \hquadric$ since for all $\yvec \in \Ucal_\xvec$, there exists a geodesic from $\xvec$ to $\yvec$ such that $\log_{\xvec} (\yvec) = \gamma_{\xvec \to \log_{\xvec} (\yvec)}'(0)$.
We show in the supp. material that the logarithm map is not defined if $ \innerlorentz{\xvec}{\yvec} \geq | \beta |$.

\textbf{Proposed dissmilarity:} We define our dissimilarity function based on the general notion of arc length and radius function on pseudo-Riemannian manifolds that we recall in the next paragraph  (see details in Chapter 5 of \cite{o1983semi}). 
This corresponds to the geodesic distance in the Riemannian case.

Let $\Ucal_\xvec$ be a normal neighborhood of $\xvec \in \Mcal$ with $\Mcal$ pseudo-Riemannian. The \textit{radius function} $r_{\xvec} : \Ucal_\xvec \to \RR$ is defined as $r_{\xvec} (\yvec) = \sqrt{\left| g_{\xvec}\left( \log_\xvec (\yvec), \log_\xvec (\yvec)\right) \right|}$ where $g_{\xvec}$ is the metric at $\xvec$. If $\sigma_{\xvec \to \xii}$ is the radial geodesic from $\xvec$ to $\yvec \in \Ucal_\xvec$ (\ie $\xii = \log_\xvec (\yvec)$), then the arc length of $\sigma_{\xvec \to \xii}$ equals $r_{\xvec} (\yvec)$.

We then define the geodesic ``distance'' between $\xvec \in \hquadric$ and $\yvec \in \Ucal_\xvec$ as the arc length of $\sigma_{\xvec \to \log_\xvec (\yvec)}$:
\begin{equation} 
\label{eq:norm_of_log_map}
 \geodistance{\xvec}{\yvec} = \sqrt{|\sqnorm{ \log_\xvec (\yvec) } |} = \left\{ 
\begin{array}{l l}
 \sqrt{| \beta |} \cosh^{-1}\left( \frac{ \langle \xvec , \yvec \rangle_q }{ \beta }\right) & \quad \text{ if }   \frac{ \langle \xvec, \yvec \rangle_q}{ | \beta |} < -1 \\ 
 0 & \quad \text{ if }  \frac{ \langle \xvec, \yvec \rangle_q }{ | \beta |} = -1 \\ 
 \sqrt{| \beta |} \cos^{-1}\left(\frac{\langle \xvec , \yvec \rangle_q}{ \beta }\right) & \quad \text{ if } \frac{ \langle \xvec, \yvec \rangle_q }{ | \beta |} \in (-1,1) %\\
 %  \sqrt{ | \beta| } \pi & \quad \text{ if } \yvec = -\xvec 
 \end{array} \right.
\end{equation}

It is important to note that our ``distance'' is {\bf not} a distance metric. However, it satisfies the axioms of a {\em symmetric premetric}: (i) $\geodistance{\xvec}{\yvec} = \geodistance{\yvec}{\xvec} \geq 0$ and (ii) $\geodistance{\xvec}{\xvec} = 0$. These conditions are sufficient to quantify the notion of nearness via a $\rho$-ball centered at $\xvec$: $B_{\xvec}^{\rho} = \left\{ \yvec : \geodistance{\xvec}{\yvec} < \rho \right\}$.

In general, topological spaces provide a qualitative (not necessarily quantitative) way to detect ``nearness'' through the concept of a neighborhood at a point \cite{lee2010introduction}. Something is true ``near $\xvec$'' if it is true in the neighborhood of $\xvec$ (\eg in $B_{\xvec}^{\rho}$). Our premetric is similar to metric learning methods \cite{7780793,8100113,xing2003distance} that learn a Mahalanobis-like distance pseudo-metric parameterized by a positive semi-definite matrix. Pairs of distinct points can have zero ``distance'' if the matrix is not positive definite.
However, unlike classic metric learning, we can have triplets ($\xvec, \yvec, \zvec$) that satisfy $\geodistance{\xvec}{\yvec} = \geodistance{\xvec}{\zvec} = 0$ but $\geodistance{\yvec}{\zvec} > 0$  (\eg $\xvec = (1,0,0,0)^{\top}, \yvec = (1,1,1,0)^{\top}, \zvec = (1,1,0,1)^{\top}$ in $\hquadricpq{2}{1}{-1}$).

Since the logarithm map is not defined if $\innerlorentz{\xvec}{\yvec} \geq | \beta |$, we propose to use the following continuous approximation defined on the whole manifold instead:

\begin{equation} \label{eq:proposed_distance}
  \forall \xvec \in \hquadric, \yvec \in \hquadric, ~ \ourdistance{\xvec}{\yvec} = 
\begin{cases}
  \geodistance{\xvec}{\yvec} & \text{ if } \innerlorentz{\xvec}{\yvec} \leq 0 \\
 \sqrt{| \beta |} \left( \frac{\pi}{2} + \frac{\innerlorentz{\xvec}{\yvec}}{|\beta|}\right) & \text{otherwise}
\end{cases}
\end{equation}

To the best of our knowledge, the explicit formulation of the logarithm map for $\hquadric$ in Eq. \eqref{eq:lambda_factor} and its corresponding radius function in Eq. \eqref{eq:norm_of_log_map} to define a dissimilarity function are novel. We have also proposed some linear approximation to evaluate dissimilarity when the logarithm map is not defined but other choices are possible. For instance, when a geodesic does not exist, a standard way in differential geometry to calculate curves is to consider broken geodesics. One might consider instead the dissimilarity $\geodistance{\xvec}{-\xvec} + \geodistance{-\xvec}{\yvec} = \pi \sqrt{| \beta |} + \geodistance{-\xvec}{\yvec}$ if $\log_\xvec (\yvec)$ is not defined  since $-\xvec \in \hquadric$ and $\log_{-\xvec} (\yvec)$ is defined. 
 This interesting problem is left for future research.

\section{Ultrahyperbolic Optimization} \label{sec:model}

In this section we present optimization frameworks to optimize any differentiable function defined on $\hquadric$. 
Our goal is to compute descent directions on the ultrahyperbolic manifold. We consider two approaches. In the first approach, we map our representation from Euclidean space to ultrahyperbolic space. This is similar to the approach taken by \cite{pmlr-v97-law19a} in hyperbolic space. In the second approach, we optimize using gradients defined directly in pseudo-Riemannian tangent space. We propose a novel descent direction which guarantees the minimization of some cost function.

\subsection{Euclidean optimization via a differentiable mapping onto $\hquadric$} \label{sec:diffeomorphisms}

Our first method maps Euclidean representations that lie in $\RRR$ to the pseudo-hyperboloid $\hquadric$, and the chain rule is exploited to perform standard gradient descent. To this end, we construct a differentiable mapping $\varphi : \RR^{q+1}_* \times \RR^p \to \hquadric$. The image of a point already on $\hquadric$ under the mapping $\varphi$ is itself: $\forall \xvec \in \hquadric, \varphi(\xvec)=\xvec$.
Let $\Scal^{q} = \left\{ \xvec \in \RR^{q+1} : \| \xvec \| = 1 \right\}$ denote the unit $q$-sphere. We first introduce the following diffeomorphisms:

\begin{theorem}[Diffeomorphisms] \label{theo:diffeomorphism}
For any $\beta < 0$, there is a diffeomorphism $\psi : \Qcal^{p,q}_\beta \to \Scal^{q} \times \RR^{p}$. Let us note $\xvec = \begin{pmatrix} \tvec \\ \svec \end{pmatrix} \in \Qcal^{p,q}_\beta$ with $\tvec \in \RR^{q+1}_*$ and $\svec \in \RR^p$, let us note $\zvec = \begin{pmatrix} \uvec \\ \vvec \end{pmatrix} \in \Scal^{q} \times \RR^{p}$ where $\uvec \in \Scal^{q}$ and $\vvec \in \RR^{p}$. The mapping $\psi$ and its inverse $\psi^{-1}$ are formulated (see proofs in supp. material):
\begin{equation}
        \psi(\xvec) = \begin{pmatrix}\frac{1}{\| \tvec \|} \tvec \\ \frac{1}{\sqrt{| \beta |}}\svec \end{pmatrix} ~~~~~~~ \text{ and } ~~~~~~ \psi^{-1}(\zvec) = \sqrt{| \beta |} \begin{pmatrix} \sqrt{1 + \| \vvec \|^2} \uvec \\ \vvec \end{pmatrix}.
\end{equation}
\end{theorem}

With these mappings, any vector $\xvec \in \RR^{q+1}_* \times \RR^p$ can be mapped to $\Qcal^{p,q}_\beta$ via $\varphi = \psi^{-1} \circ \psi$. 
$\varphi$ is differentiable everywhere except when $x_0=\cdots =x_q=0$, which should never occur in practice. It can therefore be optimized using standard gradient methods.

\subsection{Pseudo-Riemannian optimization} \label{sec:pseudo_riemannian_optimization}

We now introduce a novel method to optimize any differentiable function $f : \hquadric \to \RR$ defined on the pseudo-hyperboloid. As we show below, the (negative of the) pseudo-Riemannian gradient is not a descent direction. We propose a simple and efficient way to calculate a descent direction.

\textbf{Pseudo-Riemannian gradient:} Since $\xvec\in\hquadric$ also lies in the Euclidean ambient space $\RRR$, the function $f$ has a well defined Euclidean gradient $\eucg=\left( \partial f(\xvec)/\partial x_0, \cdots, \partial f(\xvec)/\partial x_{p+q}\right)^{\top} \in \RR^d$. 
The gradient of $f$ in the pseudo-Euclidean ambient space $ \RRpseudo$ is $(\srbasis^{-1} \eucg) = (\srbasis \eucg) \in \RRpseudo$. Since $\hquadric$ is a submanifold of $\RRpseudo$, the pseudo-Riemannian gradient $\srgrad \in \tquadric{\xvec}$ of $f$ on $\hquadric$ is the orthogonal projection of $(\srbasis \eucg)$ onto $\tquadric{\xvec}$ (see Chapter 4 of \cite{o1983semi}): 
\begin{equation} \label{eq:pseudo-gradient}
    \srgrad =  \orthoproj \left( \srbasis \eucg \right) = \srbasis \eucg - \frac{\langle \srbasis \eucg, \xvec \rangle_q}{\langle \xvec , \xvec \rangle_q} \xvec = \srbasis \eucg - \frac{\langle \eucg, \xvec \rangle}{\langle \xvec , \xvec \rangle_q} \xvec.
\end{equation} 
This gradient forms the foundation of our descent method optimizer as will be shown in Eq. \eqref{eq:taylor_approximation}.

\textbf{Iterative optimization:} Our goal is to iteratively decrease the value of the function $f$ by following some descent direction. Since $\hquadric$ is not a vector space, we do not ``follow the descent direction'' by adding the descent direction multiplied by a step size as this would result in a new point that does not necessarily lie on $\hquadric$. Instead, to remain on the manifold, we use our exponential map defined in Eq.~\eqref{eq:exponential}. This is a standard way to optimize on Riemannian manifolds \cite{absil2009optimization}. Given a step size $t>0$, one step of descent along a tangent vector $\Yvec\in\tquadric{\xvec}$ is given by:
\begin{equation}
    \yvec = \exp_{\xvec}\left(t\Yvec\right) \in \hquadric.
\end{equation}
\textbf{Descent direction:} We now explain why the negative of the pseudo-Riemannian gradient is not a descent direction. 
Our explanation extends Chapter 3 of \cite{nocedal2006numerical} that gives the criteria for a tangent vector $\zetaa$ to be a descent direction when the domain of the optimized function is a Euclidean space. 
By using the properties described in Section \ref{sec:pseudosphere}, we know that for all $t \in \RR$ and all $\xii \in \tquadric{\xvec}$, we have the equalities: $\exp_{\xvec}\left(t\xii\right) = \gamma_{\xvec \to t \xii} (1) = \gamma_{\xvec \to \xii} (t)$ so we can equivalently fix $t$ to 1 and choose the scale of $\xii$ appropriately. By exploiting Taylor's first-order approximation, there exists some small enough tangent vector $\zetaa \neq \zeroo$ (\ie with $\exp_{\xvec} (\zetaa)$ belonging to a convex neighborhood of $\xvec$ \cite{carmo1992riemannian,gao2018semi}) that satisfies the following conditions: $\gamma_{\xvec \to \zetaa} (0) = \xvec \in \hquadric$, $\gamma_{\xvec \to \zetaa}' (0) = \zetaa \in \tquadric{\xvec}$,  $\gamma_{\xvec \to \zetaa} (1) = \yvec \in \hquadric$, and the function $f \circ \gamma_{\xvec \to \zetaa} : \RR \to \RR$ can be approximated at $t = 1$ by:
\begin{equation} \label{eq:taylor_approximation}
    f ( \yvec ) = f \circ \gamma_{\xvec \to \zetaa} (1)  \simeq  f \circ \gamma_{\xvec \to \zetaa} (0) + (f \circ \gamma_{\xvec \to \zetaa})' (0) = f (\xvec) + \innerlorentz{\srgrad}{\zetaa}.
\end{equation}
where we use the following properties: $\forall t, (f \circ \gamma)' (t) = \differential(\gamma'(t)) = g_{\gamma(t)} \left( D f (\gamma(t)),\gamma'(t) \right)$ (see details in pages 11, 15 and 85 of \cite{o1983semi}), $\differential$ is the differential of $f$ and $\gamma$ is a geodesic.

To be a descent direction at $\xvec$ (\ie so that $f ( \yvec ) < f (\xvec)$), the search direction $\zetaa$ has to satisfy $\innerlorentz{\srgrad}{\zetaa} < 0$. 
However, choosing $\zetaa = - \eta \srgrad$, where $\eta > 0$ is a step size, might increase the function value if the scalar product $\innerlorentz{\cdot}{\cdot}$ is not positive definite. 
If $p + q \geq 1$, then $\innerlorentz{\cdot}{\cdot}$ is positive definite only if $q=0$ (see details in supp. material), and it is negative definite iff $p=0$ since $\innerlorentz{\cdot}{\cdot} = - \langle \cdot, \cdot \rangle$ in this case. A simple solution would be to choose $\zetaa = \pm \eta \srgrad$ depending on the sign of $\innerlorentz{\srgrad}{\zetaa}$, but $\innerlorentz{\srgrad}{\zetaa}$ might be equal to $0$ even if $\srgrad \neq \zeroo$ if $\innerlorentz{\cdot}{\cdot}$ is indefinite. The optimization algorithm might then be stuck to a level set of $f$, which is problematic.

\textbf{Proposed solution:} To ensure that $\zetaa \in \tquadric{\xvec}$ is a descent direction, we propose a simple expression that satisfies $\innerlorentz{\srgrad}{\zetaa} < 0$ if $\srgrad \neq \zeroo$ and $\innerlorentz{\srgrad}{\zetaa} = 0$ otherwise. We propose to formulate $\zetaa = - \eta \orthoproj(\srbasis \srgrad)  \in \tquadric{\xvec}$, and we define the following tangent vector $\descent = - \frac{1}{\eta} \zetaa$:
\begin{equation} \label{eq:descent}
   \descent = \orthoproj(\srbasis \srgrad) = \eucg - \frac{\inner{\eucg}{\xvec}}{\innerlorentz{\xvec}{\xvec}} \srbasis \xvec- \frac{\innerlorentz{\eucg}{\xvec}}{\innerlorentz{\xvec}{\xvec}} \xvec + \frac{\| \xvec \|^2 \inner{\eucg}{\xvec}}{\innerlorentz{\xvec}{\xvec}^2} \xvec.
\end{equation}
The tangent vector $\zetaa$ is a descent direction because $\innerlorentz{\srgrad}{\zetaa} = - \eta \innerlorentz{\srgrad}{\descent}$ is nonpositive:
\begin{align}
    \innerlorentz{\srgrad}{\descent} & = \| \eucg \|^2 - 2 \frac{\inner{\eucg}{\xvec} \innerlorentz{\eucg}{\xvec} }{\innerlorentz{\xvec}{\xvec}} + \frac{\inner{\eucg}{\xvec}^2 \| \xvec \|^2}{\innerlorentz{\xvec}{\xvec}^2} \\ & = \| \srbasis \eucg - \frac{\inner{\eucg}{\xvec}}{\innerlorentz{\xvec}{\xvec}} \xvec\|^2 = \| \srgrad \|^2 \geq 0.
\end{align}
We also have $\innerlorentz{\srgrad}{\descent} = \| \srgrad \|^2 = 0$ iff $\srgrad = \zeroo$ (\ie $\xvec$ is a stationary point). 
It is worth noting that $\srgrad = \zeroo$ implies $\descent = \orthoproj(\srbasis \zeroo) = \zeroo$. Moreover, $\descent = \zeroo$ implies that $\| \srgrad \|^2 = \innerlorentz{\srgrad}{\zeroo} = 0$.  We then have $\descent = \zeroo$ iff $\srgrad = \zeroo$.

Our proposed algorithm to the minimization problem $\min_{\xvec \in \hquadric} f(\xvec)$ is illustrated in Algorithm \ref{alg:algorithm-label}. Following generic Riemannian optimization algorithms \cite{absil2009optimization}, at each iteration, it first computes the descent direction $-\chii \in \tquadric{\xvec}$, then 
decreases the function by applying the exponential map defined in Eq.~\eqref{eq:exponential}. 
It is worth noting that our proposed descent method can be applied to any differentiable function $f : \hquadric \to \RR$, not only to those that exploit the distance introduced in Section~\ref{sec:pseudosphere}.

\begin{algorithm}[!t]
    \caption{Pseudo-Riemannian optimization on $\hquadric$}
    \label{alg:algorithm-label}
    \footnotesize
\begin{flushleft}
        \textbf{input:} differentiable function $f : \hquadric \to \RR$ to be minimized, some initial value of $\xvec \in \hquadric$ 
\end{flushleft}    
        \begin{algorithmic}[1]
    \While{not converge}
        \State Calculate  $\eucg$  \Comment{\ie the Euclidean gradient of $f$ at $\xvec$ in the Euclidean ambient space}
        \State $\descent \leftarrow \orthoproj(\srbasis \orthoproj(\srbasis \eucg))$  \Comment{see Eq. \eqref{eq:descent}}
        \State $\xvec \leftarrow \exp_{\xvec} (- \eta \descent)$ \Comment{where $\eta > 0$ is a step size (\eg determined with line search)}
    \EndWhile
    \end{algorithmic}
\end{algorithm}

Interestingly, our method can also be seen as a preconditioning technique \cite{nocedal2006numerical} where the descent direction is obtained by preconditioning the pseudo-Riemannian gradient $\srgrad$ with the matrix $\preconditioner =\left[\srbasis - \frac{1}{\innerlorentz{\xvec}{\xvec}} \xvec \xvec^{\top} \right] \in \RR^{d \times d}$. 
In other words, we have $\descent = \preconditioner \srgrad = \orthoproj(\srbasis \srgrad)$.

In the more general setting of pseudo-Riemannian manifolds, another preconditioning technique was proposed in \cite{gao2018semi}. The method in \cite{gao2018semi} requires performing a Gram-Schmidt process at each iteration to obtain an (\textit{ordered} \cite{wolf1972spaces}) orthonormal basis of the tangent space at $\xvec$ \wrt the induced quadratic form of the manifold. However, the Gram-Schmidt process is unstable and has algorithmic complexity that is cubic in the dimensionality of the tangent space. On the other hand, our method is more stable and its algorithmic complexity is linear in the dimensionality of the tangent space.

\section{Experiments}

We now experimentally validate our proposed optimization methods and the effectiveness of our dissimilarity function. Our main experimental results can be summarized as follows: 

$\bullet$ Both optimizers introduced in Section \ref{sec:model} decrease some objective function $f : \hquadric \to \RR$.
While both optimizers manage to learn high-dimensional representations that satisfy the problem-dependent training constraints, only the pseudo-Riemannian optimizer satisfies all the constraints in lower-dimensional spaces. This is because it exploits the underlying metric of the manifold.

$\bullet$ Hyperbolic representations are popular in machine learning as they are well suited to represent hierarchical trees \cite{gromov1987hyperbolic,nickel2017poincare,nickel2018learning}. 
On the other hand, hierarchical datasets whose graph contains cycles cannot be represented using trees. Therefore, we propose to represent such graphs using our ultrahyperbolic representations. 
An important example are community graphs such as Zachary's karate club \cite{zachary1977information} that contain leaders. Because our ultrahyperbolic representations are more flexible than hyperbolic representations, we believe that our representations are better suited for these non tree-like hierarchical structures.

\textbf{Graph:} Our ultrahyperbolic representations describe graph-structured datasets. Each dataset is an undirected weighted graph $\graph = (\nodes, \edges)$ which has node-set $\nodes = \{ v_i \}_{i=1}^{n}$ and edge-set $\edges = \{ e_k \}_{k=1}^m$. Each edge $e_k$ is weighted by an arbitrary capacity $c_k \in \RR_+$ that models the strength of the relationship between nodes. The higher the capacity $c_k$, the stronger the relationship between the nodes connected by $e_k$. 

\textbf{Learned representations:} Our problem formulation is inspired by hyperbolic representation learning approaches \cite{nickel2017poincare,nickel2018learning} where the nodes of a tree (\ie graph without cycles) are represented in hyperbolic space. The hierarchical structure of the tree is then reflected by the order of distances between its nodes. More precisely, a node representation is learned so that each node is closer to its descendants and ancestors in the tree (\wrt the hyperbolic distance) than to any other node. For example, in a hierarchy of words, ancestors and descendants are hypernyms and hyponyms, respectively.

\begin{figure}[!t]
    \centering
    \includegraphics[width=.355\linewidth]{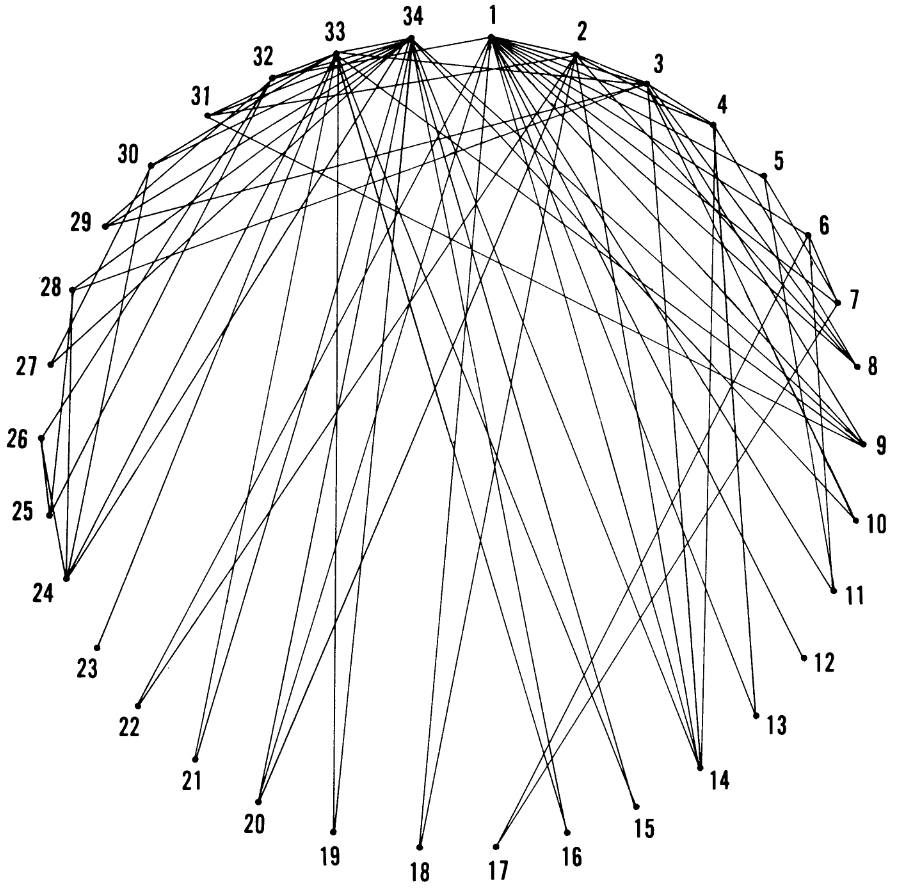}~~\includegraphics[width=.62\linewidth]{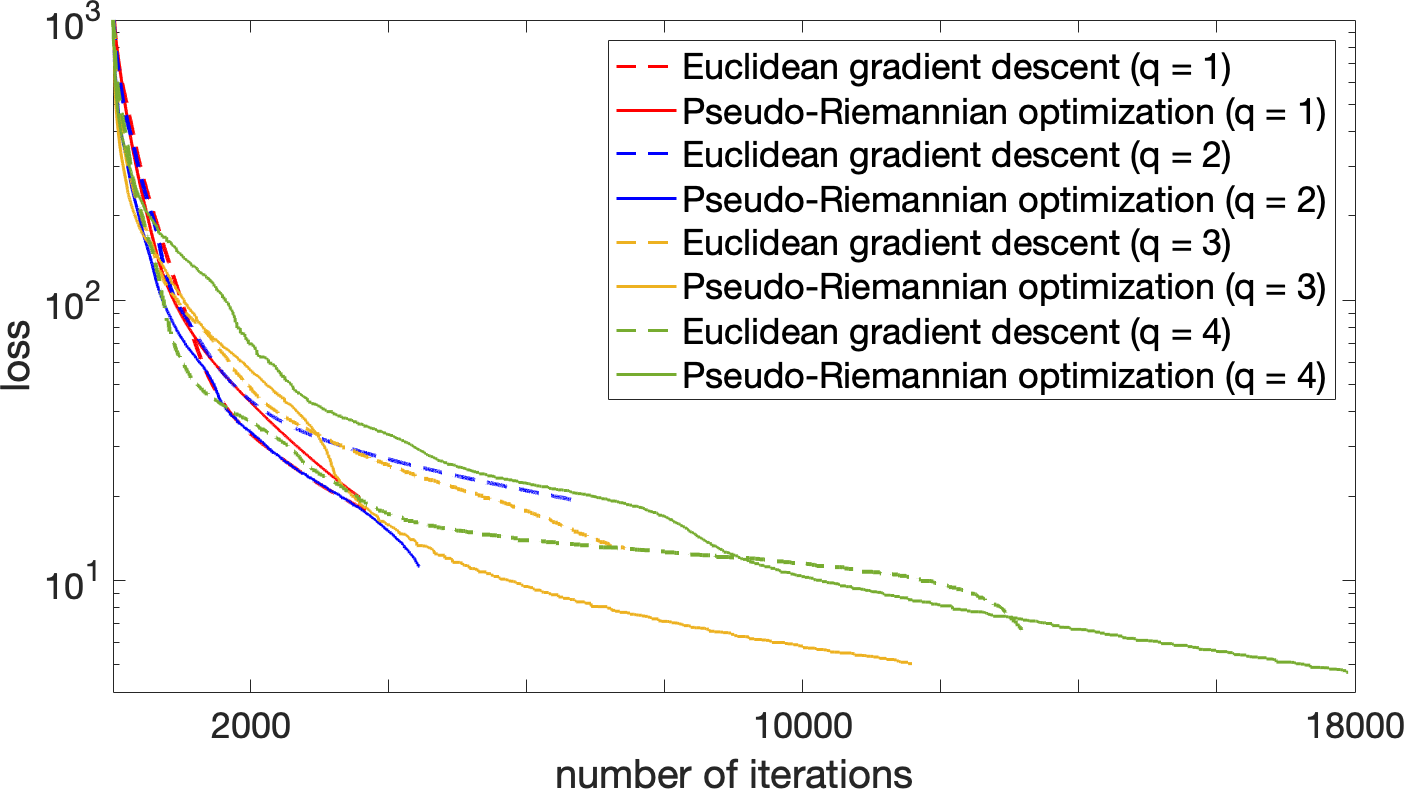}
    \caption{(left) graphic representation of Zachary's karate club (figure courtesy of \cite{zachary1977information}). (right) Loss values of Eq. \eqref{eq:problem} for different numbers of time dimensions and optimizers.}
    \label{fig:zachary}
\end{figure}

Our goal is to learn a set of $n$ points $\xvec_1, \cdots, \xvec_n \in \hquadric$ (embeddings) from a given graph $\graph$. 
The presence of cycles in the graph makes it difficult to determine ancestors and descendants. For this reason, we introduce for each pair of nodes $(v_i, v_j) = e_k \in E$, the set of ``\textit{weaker}'' pairs that have lower capacity: $\Wcal(e_k) = \left\{ e_l : c_k > c_l\right\} \cup \left\{ (v_a, v_b) : (v_a, v_b) \notin E \right\}$. 
Our goal is to learn representations such that pairs $(v_i, v_j)$ with higher capacity have their representations $(\xvec_i,\xvec_j)$ closer to each other than weaker pairs. Following \cite{nickel2017poincare}, we formulate our problem as:
\begin{equation} \label{eq:problem}
    \min_{\xvec_1, \cdots, \xvec_n \in \hquadric} \sum_{(v_i, v_j) ~ = ~ e_k \in \edges} - \log \frac{\exp\left(-\dsf(\xvec_i,\xvec_j)/\tau \right)}{ \displaystyle \sum_{(v_a, v_b) \in ~ \Wcal(e_k) \cup \{ e_k \}} \exp\left(-\dsf(\xvec_a,\xvec_b)/\tau \right)}
\end{equation}
where $\dsf$ is the chosen dissimilarity function (\eg $\ourdistance{\cdot}{\cdot}$ defined in Eq. \eqref{eq:proposed_distance}) and $\tau > 0$ is a fixed temperature parameter. The formulation of Eq. \eqref{eq:problem} is classic in the metric learning literature \cite{Cao2020A,law2018dimensionality,8683393} and corresponds to optimizing some order on the learned distances via a softmax function.

\textbf{Implementation details:} We coded our approach in PyTorch \cite{NEURIPS2019_9015} that automatically calculates the Euclidean gradient $\eucgi$.
Initially, a random set of vectors $\left\{\zvec_i\right\}_{i=1}^{n}$ is generated close to the positive pole $(\sqrt{| \beta |}, 0, \cdots, 0) \in \hquadric$ with every coordinate perturbed uniformly with a random value in the interval $[-\varepsilon,\varepsilon]$ where $\varepsilon > 0$ is chosen small enough so that $\sqnorm{\zvec_i} < 0$. We set $\beta = -1$, $\varepsilon = 0.1$ and $\tau = 10^{-2}$. Initial embeddings are generated as follows: $\forall i, \xvec_i = \sqrt{| \beta |} \frac{\zvec_i}{\sqrt{|\sqnorm{\zvec_i}|}} \in \hquadric$.

\textbf{Zachary's karate club dataset} \cite{zachary1977information} is a social network graph of a karate club comprised of $n = 34$ nodes, each representing a member of the karate club. The club was split due to a conflict between instructor "Mr. Hi" (node $v_{1}$) and administrator "John A" (node $v_{n}$). The remaining members now have to decide whether to join the new club created by $v_1$ or not. 
In \cite{zachary1977information}, Zachary defines a matrix of relative strengths of the friendships in the karate club called $\Capacity \in \{ 0, 1, \cdots ,7\}^{n \times n}$ and that depends on various criteria. We note that the matrix is not symmetric and has 7 different pairs $(v_i,v_j)$ for which $\Capacity_{ij} \neq \Capacity_{ji}$. Since our dissimilarity function is symmetric, we consider the symmetric matrix $\SymCapacity = \Capacity + \Capacity^{\top}$ instead. The value of $\SymCapacity_{ij}$ is the capacity/weight assigned to the edge joining $v_i$ and $v_j$, and there is no edge between $v_i$ and $v_j$ if $\SymCapacity_{ij} = 0$. Fig. \ref{fig:zachary} (left) illustrates the 34 nodes of the dataset, an edge joining the nodes $v_i$ and $v_j$ is drawn iff $\SymCapacity_{ij} \neq 0$. The level of a node in the hierarchy corresponds approximately to its height in the figure. 

\textbf{Optimizers:} We validate that our optimizers introduced in Section \ref{sec:model} decrease the cost function. 
First, we consider the simple unweighted case where every edge weight is 1. 
For each edge $e_k \in \edges$, $\Wcal(e_k)$ is then the set of pairs of nodes that are not connected. In other words, 
Eq. \eqref{eq:problem} learns node representations that have the property that every connected pair of nodes has smaller distance than non-connected pairs. We use this condition as a stopping criterion of our algorithm. 

Fig. \ref{fig:zachary} (right) illustrates the loss values of Eq. \eqref{eq:problem} as a function of the number of iterations with the Euclidean gradient descent (Section \ref{sec:diffeomorphisms}) and our pseudo-Riemannian optimizer (introduced in Section \ref{sec:pseudo_riemannian_optimization}).
In each test, we vary the number of time dimensions $q+1$ while the ambient space is of fixed dimensionality $d= p + q + 1 = 10$. We omit the case $q=0$ since it corresponds to the (hyperbolic) Riemannian case already considered in \cite{pmlr-v97-law19a,nickel2018learning}. Both optimizers decrease the function and manage to satisfy all the expected distance relations. 
We note that when we use $-\srgrad$ instead of $-\chii$ as a search direction, the algorithm does not converge. 
Moreover, our pseudo-Riemannian optimizer manages to learn representations that satisfy all the constraints for low-dimensional manifolds such as $\Qcal_{-1}^{4,1}$ and $\Qcal_{-1}^{4,2}$, while the optimizer introduced in Section \ref{sec:diffeomorphisms} does not. 
Consequently, we only use the pseudo-Riemannian optimizer in the following results.

\begin{table}[!t]
  \caption{Evaluation scores for the different learned representations (mean $\pm$ standard deviation)}
  \label{table:scores}
  \centering \scriptsize 
  \begin{tabular}{lcccccc}
    \toprule
    Evaluation metric   & $\RR^4$     & $\Qcal^{4,0}_{-1}$ & $\Qcal^{3,1}_{-1}$ & $\Qcal^{2,2}_{-1}$  & $\Qcal^{1,3}_{-1}$ & $\Qcal^{0,4}_{-1}$ \\ & (flat) & (hyperbolic) & (ours) & (ours) & (ours) & (spherical) \\ \midrule 
    %Evaluation metric   & Euclidean     & $q = 0$ (hyperbolic) & $q = 1$ (ours) & $q = 2$ (ours) & $q = 3$ (ours)\\ \midrule
    Rank of the first leader & $5.4 \pm 1.1$ & $2.8 \pm 0.4$ & $\textbf{1.2} \pm \textbf{0.4}$ & $\textbf{1.2} \pm \textbf{0.4}$ & $ 1.8 \pm 0.8$ & $2.0 \pm 0.7$ \\
    Rank of the second leader   & $6.6 \pm 0.9$  & $4.2 \pm 0.7$ & $\textbf{2.4} \pm \textbf{0.9}$ & $2.6 \pm 0.5$   & $4.0 \pm 1.2$  & $4.0 \pm 1.4$ \\
      top 5 Spearman's $\rho$ & $-0.44 \pm 0.19$ & $0.20 \pm 0.48$& $\textbf{0.76} \pm \textbf{0.21}$ & $0.66 \pm 0.30$& $0.36 \pm 0.40$ & $0.18 \pm 0.37$\\ 
      top 10 Spearman's $\rho$ & $0.00 \pm 0.14$ & $0.38 \pm 0.06$   & $0.74 \pm 0.11$ & $\textbf{0.79} \pm \textbf{0.12}$ & $0.71 \pm 0.08$  & $0.55 \pm 0.20$\\
    \bottomrule
  \end{tabular}
\end{table}

\textbf{Hierarchy extraction:} To quantitatively evaluate our approach, we apply it to the problem of predicting the high-level nodes in the hierarchy from the weighted matrix $\SymCapacity$ given as supervision. We consider the challenging low-dimensional setting where all the learned representations lie on a 4-dimensional manifold (\ie $p + q + 1 = 5$). Hyperbolic distances are known to grow exponentially as we get further from the origin. Therefore, the sum of distances $\delta_i = \sum_{j=1}^n \dsf(\xvec_i, \xvec_j)$ of a node $v_i$ with all other nodes is a good indication of importance. Intuitively, high-level nodes will be closer to most nodes than low-level nodes. We then sort the scores $\delta_1, \cdots, \delta_n$ in ascending order and report the ranks of the two leaders $v_1$ or $v_n$ (in no particular order) in the first two rows of Table \ref{table:scores} averaged over 5 different initializations/runs. Leaders tend to have a smaller $\delta_i$ score with ultrahyperbolic distances than with Euclidean, hyperbolic or spherical distances. 

Instead of using $\delta_i$ for hyperbolic representations, the importance of a node $v_i$ can be evaluated by using the Euclidean norm of its embedding $\xvec_i$ as proxy \cite{pmlr-v97-law19a,nickel2017poincare,nickel2018learning}. This is because high-level nodes of a tree in hyperbolic space are usually closer to the origin than low-level nodes. 
Not surprisingly, this proxy leads to worse performance ($8.6 \pm 2.3$ and $18.6 \pm 4.9$) as the relationships are not that of a tree. 
Since hierarchy levels are hard to compare for low-level nodes, we select the 10 (or 5) most influential members based on the score $s_i = \sum_{j=1}^n \SymCapacity_{ij}$. The corresponding nodes are 34, 1, 33, 3, 2, 32, 24, 4, 9, 14 (in that order). Spearman's rank correlation coefficient \cite{spearman2015proof} between the selected scores $s_i$ and corresponding $\delta_i$ is reported in Table \ref{table:scores} and shows the relevance of our representations.

Due to lack of space, we also report in the supp. material similar experiments on a larger hierarchical dataset \citep{chechik2007eec} that describes co-authorship from papers published at NIPS from 1988 to 2003.

\section{Conclusion} \label{sec:conclusion}

We have introduced ultrahyperbolic representations. Our representations lie on a pseudo-Riemannian manifold of constant nonzero curvature which generalizes hyperbolic and spherical geometries and includes them as submanifolds. Any relationship described in those geometries can then be described with our representations that are more flexible. We have introduced new optimization tools and experimentally shown that our representations can extract hierarchies in graphs that contain cycles.

\section*{Broader Impact}

We introduce a novel way of representing relationships between data points by considering the geometry of non-Riemannian manifolds of constant nonzero curvature. The relationships between data points are described by a dissimilarity function that we introduce and exploits the structure of the manifold. It is more flexible than the distance metric used in hyperbolic and spherical geometries often used in machine learning and computer vision. Nonetheless, since the problems involving our representations are not straightforward to optimize, we propose novel optimization algorithms that can potentially benefit the machine learning, computer vision and natural language processing communities. Indeed, our method is application agnostic and could extend existing frameworks.

Our contribution is mainly theoretical but we have included one practical application. Similarly to hyperbolic representations that are popular for representing tree-like data, we have shown that our representations are well adapted to the more general case of hierarchical graphs with cycles. These graphs appear in many different fields of research such as medicine, molecular biology and the social sciences. For example, an ultrahyperbolic representation of proteins might assist in understanding their complicated folding mechanisms. Moreover, these representations could assist in analyzing features of social media such as discovering new trends and leading "connectors". The impact of community detection for commercial or political advertising is already known in social networking services. We foresee that our method will have many more graph-based practical applications.

We know of very few applications outside of general relativity that use pseudo-Riemannian geometry. We hope that our research will stimulate other applications in machine learning and related fields. 
Finally, although we have introduced a novel descent direction for our optimization algorithm, future research could study and improve its rate of convergence. 

\begin{ack}
We thank Jonah Philion, Guojun Zhang and the anonymous reviewers for helpful feedback on early versions of this manuscript. 

This article was entirely funded by NVIDIA corporation. Marc Law and Jos Stam completed this working from home during the COVID-19 pandemic. 
\end{ack}
\small
\bibliography{neurips_2020}
\bibliographystyle{plain}
\normalsize

\newpage

\appendix

\section{Supplementary material of ``Ultrahyperbolic Representation Learning''}

The supplementary material is structured as follows:

$\bullet$ In Section \ref{sec:geometry_discussion}, we provide a short discussion on the choice of geometry to represent graphs.

$\bullet$ In Section \ref{sec:geodesic}, we study the formulation of the geodesics in Eq. \eqref{eq:geodesic}. It is combination of hyperbolic, spherical and flat space due to its formulation.

$\bullet$ In Section \ref{sec:neighborhood}, we explain why $\log_\xvec (\yvec)$ (see Eq. \eqref{eq:lambda_factor}) is not defined if $\innerlorentz{\xvec}{\yvec} \geq | \beta |$. 

$\bullet$ In Section \ref{sec:antiisometry}, we explain the anti-isometry between $\Qcal^{p,q}_\beta$ or $\Qcal^{q+1,p-1}_{-\beta}$.

$\bullet$ In Section \ref{sec:curvature}, we study the curvature of $\Qcal^{p,q}_\beta$.

$\bullet$ In Section \ref{sec:diffeomorphism_proof}, we give the proof of Theorem \ref{theo:diffeomorphism}. 

$\bullet$ In Section \ref{sec:hyperboloid}, we explain why the upper sheet of the two-sheet hyperboloid $\hquadricpq{p}{0}{\beta}$ (\ie the case where $q = 0$) is a Riemannian manifold even if its metric matrix $\srbasis = \textbf{I}_{1,p}$ is not positive definite.

$\bullet$ In Section 
\ref{sec:experiments_supp}, we report additional experiments.

\section{Choice of Geometry} \label{sec:geometry_discussion}

The choice of geometry to represent graphs is still an open problem in general. It depends on the topology of the graph and the kind of relationship between nodes. For instance, hyperbolic geometry was mathematically shown to be appropriate for tree-like graphs \cite{gromov1987hyperbolic}, but not for other types of graphs. Ultrahyperbolic geometry has the advantage of generalizing both hyperbolic and spherical geometries and can describe relationships specific to those geometries. In particular, the geodesic \textit{distance} can be written in the same way as the \Poincare and spherical distances as shown in Eq. \eqref{eq:norm_of_log_map},  Some parts of the manifold are hyperbolic or spherical as explained in the paper. The converse is not true. The framework might then automatically learn representations to be part of a same hyperbolic or spherical part of the manifold depending on the context. 

Those reasons led us to consider hierarchical graphs that were similar to trees, but where the presence of cycles in the graph limited the relevance of hyperbolic geometry. We experimentally validated our intuition. 
The choice of manifold of constant nonzero curvature (i.e. the optimal number of time and space dimensions $q + 1$ and $p$) then seems to depend on how much the graph is similar to a tree or to a graph where spherical geometry is appropriate, such as cycle graphs (or a mix of both). 
%Nonetheless, determining the optimal number of time and space dimensions (i.e. $q$ and $p$) is still an open problem. 

It is also worth noting that ultrahyperbolic geometry can describe some graph concepts differently, if not better, than hyperbolic and spherical geometries. 
For instance, it is known that \textit{triadic closure} is a concept in social network theory that is not valid for most large and complex networks. In other words, if $(\xvec, \yvec)$ and $(\xvec, \zvec)$ are strongly tied, triadic closure would imply that $(\yvec, \zvec)$ are strongly tied as well. 
As explained in Section \ref{sec:pseudosphere}, the fact that we can find triplets that satisfy $\geodistance{\xvec}{\yvec} = \geodistance{\xvec}{\zvec} = 0$ but $\geodistance{\yvec}{\zvec} > 0$ allows to avoid triadic closure. 
Moreover, although we have applied our proposed representations to some type of graph, they can be applied to other applications that do not involve graphs.  The possible applications are left for future research. 

\section{Some properties about geodesics and logarithm map} \label{sec:differential_geometry}

\subsection{Geodesics of $\hquadric$} \label{sec:geodesic}

\textbf{Tangent space of pseudo-Riemannian submanifolds:} 
In the paper, we exploit the fact that our considered manifold $\Mcal$ (here $\hquadric$) is a pseudo-Riemannian submanifold of $\MMcal$ (here $\RRpseudo$). 
Since $\MMcal$ is chosen to be a vector space, we have a natural isomorphism between $\MMcal$ and its tangent space. 

If $\Mcal$ is a pseudo-Riemannian submanifold of $\MMcal$, we have the following direct sum decomposition:
\begin{equation}
    T_{\xvec} (\MMcal) = T_{\xvec} (\Mcal) + T_{\xvec} (\Mcal)^{\bot},
\end{equation}
where $T_{\xvec} (\Mcal)^{\bot} = \{ \zetaa \in T_{\xvec} (\MMcal) : \forall \xii \in T_{\xvec} (\Mcal), ~  \overbar{g_{\xvec}}(\zetaa,\xii) = 0 \}$ is the orthogonal complement of $T_{\xvec} (\Mcal)$ and called the \textit{normal space} of $\Mcal$ at $\xvec$. It is a nondegenerate subspace of $T_{\xvec} (\MMcal)$, and $\overbar{g_{\xvec}}$ is the metric at $\xvec \in \MMcal$. In the case of $\hquadric$, $T_{\xvec} (\hquadric)^{\bot}$ is defined as:
\begin{align}
    & T_{\xvec} (\hquadric)^{\bot} = \{ \zetaa \in T_{\xvec} \RRpseudo : \forall \xii \in T_{\xvec} \hquadric, ~ \innerlorentz{\zetaa}{\xii} = 0 \} \\
    \approx ~~ & N_{\xvec} (\hquadric, \RRpseudo) =  \{ \lambda \xvec \in \RRpseudo : \lambda \in \RR \}.
\end{align}
where $\approx$ denotes isomorphism.

\textbf{Geodesic of a submanifold:} As mentioned in the main paper, a curve $\gamma$ is a geodesic if its acceleration is zero. However, the acceleration depends on the choice of the affine connection while the velocity does not (\ie the velocity does not depend on the Christoffel symbols whereas the acceleration does, and different connections produce different geodesics, see details in page 66 of \cite{o1983semi} or Chapter 5.4 of \cite{absil2009optimization}). Let us note $\frac{\overbar{\covariantderivative}}{dt} \left(\gamma'(t) \right)$ (resp. $\frac{\covariantderivative}{dt} \left(\gamma'(t) \right)$) the covariant derivative of $\gamma'(t)$ along $\gamma(t)$ in $\RRpseudo$ (resp. in $\hquadric$). 
By using the \textit{induced connection} (see page 98 of \cite{o1983semi}) and the fact that $\hquadric$ is isometrically embedded in $\RRpseudo$, the second-order ordinary differential equation about the zero acceleration of the geodesic is equivalent to (see page 103 of \cite{o1983semi}):
\begin{equation} \label{eq:second_order}
    \frac{\overbar{\covariantderivative}}{dt} \left(\gamma'(t) \right) \in N_{\gamma (t)} (\hquadric, \RRpseudo) \iff \gamma''(t) = \frac{\covariantderivative}{dt} \left(\gamma'(t) \right) = \orthoproji{\gamma(t)} \left( \frac{\overbar{\covariantderivative}}{dt} \left(\gamma'(t) \right) \right) = \zeroo.
\end{equation}
where $\orthoproji{\gamma(t)}$ is defined in Eq. \eqref{eq:ortho_projection} and orthogonally projects onto $\tquadric{\gamma(t)}$. 
In other words, $\gamma$ is straight in $\Mcal$ but its curving in $\MMcal$ is the one forced by the curving of $\Mcal$ itself in $\MMcal$ \cite{o1983semi}. 

The initial conditions $\gamma_{\xvec \to \xii} (0) = \xvec \in \hquadric$ and $\gamma'_{\xvec \to \xii} (0) = \xii \in \tquadric{\xvec}$ are straightforward. 
With the formulation of $\gamma_{\xvec \to \xii}$ in Eq. \eqref{eq:geodesic}, 
one can first verify that for all $t$, $\gamma_{\xvec \to \xii} (t)$ lies on $\hquadric$. Indeed, since $\innerlorentz{\xvec}{\xii} = 0$, we have:
\begin{equation}
 \forall t \in \RR, ~ \forall \xvec \in \hquadric, ~ \forall \xii \in \tquadric{\xvec}, ~  \innerlorentz{\gamma_{\xvec \to \xii} (t)}{\gamma_{\xvec \to \xii} (t)} = \beta.
\end{equation}

The acceleration of $\gamma$ at $t$ (in the ambient space $\RRpseudo$) is in $N_{\gamma(t)} (\hquadric, \RRpseudo)$ since we have $\forall \xvec \in \hquadric, ~ \forall \xii \in \tquadric{\xvec}$:
\begin{align} \label{eq:geodesic_supp}
& \forall t \in \RR, ~ \frac{\overbar{\covariantderivative}}{dt} \left(\gamma'_{\xvec \to \xii}(t) \right) =
\frac{\innerlorentz{\xii}{\xii}}{| \beta |} \gamma_{\xvec \to \xii} (t) \in N_{\gamma_{\xvec \to \xii} (t)} (\hquadric, \RRpseudo) \\
\Longrightarrow & \forall t \in \RR,  ~  \gamma_{\xvec \to \xii}''(t) = \frac{\covariantderivative}{dt} \left( \gamma_{\xvec \to \xii}'(t) \right) = \zeroo. \label{eq:second_order_ode_zero}
\end{align}

We have found a solution for our second-order ordinary differential equation, the solution is unique by definition. \qed

From its formulation in Eq. \eqref{eq:geodesic}, the nonconstant geodesic $\gamma_{\xvec \to \xii}$ is similar to the hyperbolic, flat or spherical case if $\innerlorentz{\xii}{\xii}$ is positive, zero or negative, respectively.

\subsection{On the non existence of the logarithm map of some points} \label{sec:neighborhood}

We explain here why $\log_\xvec (\yvec)$ is not defined if $\innerlorentz{\xvec}{\yvec} \geq | \beta |$. 
Our proof relies on the fact the domain of the exponential map is the whole tangent space. 
We also use the fact that $\forall \xii \in T_{\xvec} \Qcal^{p,q}_{\beta},$ we have $\innerlorentz{\xvec}{\xii} = 0$ by definition of tangent vectors to $\hquadric$.

Assume that there exists a tangent vector $\xii \in T_{\xvec} \Qcal^{p,q}_{\beta}$ such that $\yvec = \exp_\xvec(\xii)$ and $\langle \yvec , \xvec \rangle_q \geq | \beta |$. We consider the three possible cases of sign of $\innerlorentz{\xii}{\xii}$.

$\bullet$ Let us first assume that $ \dinnerlorentz{\xii} > 0$. 
For all $\xii \in T_{\xvec} \Qcal^{p,q}_{\beta}$ that satisfies $\langle \xii, \xii \rangle_q > 0$, we have by definition of the corresponding exponential map:
\begin{align}
  \langle \xvec , \exp_\xvec (\xii) \rangle_q & = \sqnorm{\xvec}\cosh \left( \frac{\sqrt{| \langle \xii, \xii \rangle_q |}}{\sqrt{| \beta |}} \right) + \sqrt{| \beta |} \frac{\innerlorentz{\xvec}{\xii}}{\sqrt{| \langle \xii, \xii \rangle_q |}} \sinh \left(\frac{ \sqrt{| \langle \xii, \xii \rangle_q |}}{\sqrt{| \beta |}} \right)  \\
  & = \sqnorm{\xvec}\cosh \left( \frac{\sqrt{| \langle \xii, \xii \rangle_q |}}{\sqrt{| \beta |}} \right) < 0.
\end{align}
Therefore, there exists no tangent vector $\xii \in  T_{\xvec} \Qcal^{p,q}_{\beta}$ that satisfies $\langle \xii, \xii \rangle_q > 0$ and $\innerlorentz{\xvec}{\exp_\xvec (\xii)} > | \beta | > 0$.

$\bullet$ Let us now assume that $ \dinnerlorentz{\xii} = 0$. 
For all $\xii \in T_{\xvec} \Qcal^{p,q}_{\beta}$ that satisfies $\langle \xii, \xii \rangle_q = 0$, we have:
\begin{equation}
    \innerlorentz{\xvec}{\exp_\xvec (\xii)} = \innerlorentz{\xvec}{\xvec + \xii} = \innerlorentz{\xvec}{\xvec} + \innerlorentz{\xvec}{\xii} = \innerlorentz{\xvec}{\xvec} = \beta < 0.
\end{equation}

$\bullet$ Let us now assume that $ \dinnerlorentz{\xii} < 0$. 
For all $\xii \in T_{\xvec} \Qcal^{p,q}_{\beta}$ that satisfies $\langle \xii, \xii \rangle_q < 0$, we have:
\begin{align}
  \langle \xvec , \exp_\xvec (\xii) \rangle_q & = \sqnorm{\xvec}\cos \left( \frac{\sqrt{| \langle \xii, \xii \rangle_q |}}{\sqrt{| \beta |}} \right) + \sqrt{| \beta |} \frac{\innerlorentz{\xvec}{\xii}}{\sqrt{| \langle \xii, \xii \rangle_q |}} \sin \left(\frac{ \sqrt{| \langle \xii, \xii \rangle_q |}}{\sqrt{| \beta |}} \right)  \\
  & = \sqnorm{\xvec}\cos \left( \frac{\sqrt{| \langle \xii, \xii \rangle_q |}}{\sqrt{| \beta |}} \right) \in [\beta, |\beta|].
\end{align}
Given the formulation of the geodesic in Eq. \eqref{eq:geodesic}, $\innerlorentz{\xvec}{\yvec} =  \innerlorentz{\xvec}{\exp_\xvec (\xii)} \geq |\beta|$ iff $\yvec = -\xvec$. 

From our study above, there exists no geodesic (hence no logarithm map) joining $\xvec$ and $\yvec$ if $ \innerlorentz{\xvec}{\yvec} \geq | \beta |$ except if $\yvec = - \xvec$. 
The antipodal point $\yvec = - \xvec$ is also a special case for which there does not exist a logarithm map since $\xvec$ and $-\xvec$ are joined by infinitely many minimizing geodesics of equal length (a similar case is the $q$-sphere). 
Nonetheless, the geodesic ``distance'' between $\xvec$ and $-\xvec$ can be defined as $\pi \sqrt{| \beta |}$.
 \qed

\subsection{Anti-isometry between $\Qcal^{p,q}_\beta$ and $\Qcal^{q+1,p-1}_{-\beta}$ } \label{sec:antiisometry}

In the main paper, we state that there is an anti-isometry between $\Qcal^{p,q}_\beta$ and $\Qcal^{q+1,p-1}_{-\beta}$. We recall that $\Qcal^{q+1,p-1}_{-\beta}$ is embedded in $\mathbb{R}^{q+1,p}$. 
Let us note $\sigma : \mathbb{R}^{p,q+1} \to \mathbb{R}^{q+1,p}$ the mapping defined as:
\begin{equation}
    \forall \xvec = (x_0, \cdots, x_{p+q})^{\top} \in \mathbb{R}^{p,q+1}, ~ \sigma(\xvec) = (x_{p+q}, x_{p+q-1}, \cdots, x_1, x_0)^{\top}.
\end{equation}
We have: 
\begin{equation}
    \forall \xvec \in  \RRpseudo, \yvec \in  \RRpseudo, ~ \innerlorentzq{\xvec}{\yvec}{q} = - \innerlorentzq{\sigma(\xvec)}{\sigma(\yvec)}{p-1}.
\end{equation}

\subsection{Curvature of $\Qcal^{p,q}_\beta$} \label{sec:curvature}

The manifold $\Qcal^{p,q}_\beta$ is a total umbilic hypersurface of $\RRpseudo$ and has constant sectional curvature $1/\beta$ and constant mean curvature $\kappa = | \beta |^{-1/2}$ with respect to the unit normal vector field $\Ncal (\xvec) = -\kappa \xvec$. More details can be found in Chapter 3 of \cite{anciaux2011minimal}.

\subsection{Proof of Theorem \ref{theo:diffeomorphism}} \label{sec:diffeomorphism_proof}

We give the proofs for Theorem \ref{theo:diffeomorphism} that we recall below:

\begin{theorem}[Diffeomorphisms] 
For any $\beta < 0$, there is a diffeomorphism $\psi : \Qcal^{p,q}_\beta \to \Scal^{q} \times \RR^{p}$. Let us note $\xvec = \begin{pmatrix} \tvec \\ \svec \end{pmatrix} \in \Qcal^{p,q}_\beta$ with $\tvec \in \RR^{q+1}_*$ and $\svec \in \RR^p$, let us note $\zvec = \begin{pmatrix} \uvec \\ \vvec \end{pmatrix} \in \Scal^{q} \times \RR^{p}$ where $\uvec \in \Scal^{q}$ and $\vvec \in \RR^{p}$. The mapping $\psi$ and its inverse $\psi^{-1}$ are formulated:
\begin{equation}
        \psi(\xvec) = \begin{pmatrix}\frac{1}{\| \tvec \|} \tvec \\ \frac{1}{\sqrt{| \beta |}}\svec \end{pmatrix} ~~~~~~~ \text{ and } ~~~~~~ \psi^{-1}(\zvec) = \sqrt{| \beta |} \begin{pmatrix} \sqrt{1 + \| \vvec \|^2} \uvec \\ \vvec \end{pmatrix}.
\end{equation}
\end{theorem}

We need to show that $\psi(\psi^{-1}(\zvec)) = \zvec$ and $\psi^{-1}(\psi(\xvec)) = \xvec$. 

\textbf{Space dimensions:} The mapping of the space dimensions of $\xvec$ to the space dimensions of $\zvec$ simply involves a scaling factor of $\sqrt{| \beta |}$ (\ie $\vvec = \frac{1}{\sqrt{| \beta |}}\svec$, and $\svec = \sqrt{| \beta |} \vvec$). 

\textbf{Time dimensions:} We show the invertibility of the mappings taking time dimensions as input.

$\bullet$ Let us first show $\psi^{-1}(\psi(\xvec)) = \xvec$. 
We recall that $\xvec = \begin{pmatrix} \tvec \\ \svec \end{pmatrix} \in \Qcal^{p,q}_\beta$ with $\tvec \in \RR^{q+1}_*$ and $\svec \in \RR^p$.

We need to show that: 
\begin{equation} \label{eq:time_diffeomorphism1}
\tvec = \sqrt{| \beta |}  \sqrt{1 + \| \vvec \|^2} \frac{\tvec}{\| \tvec \|} = \sqrt{| \beta |}  \sqrt{1 + \| \frac{1}{\sqrt{| \beta |}}\svec \|^2} \frac{\tvec}{\| \tvec \|} = \sqrt{| \beta |}  \sqrt{1 + \frac{1}{| \beta |} \| \svec \|^2} \frac{\tvec}{\| \tvec \|}.
\end{equation}

To show Eq. \eqref{eq:time_diffeomorphism1}, it is sufficient to prove that the following property is satisfied:
\begin{equation}
 \| \tvec \| = \sqrt{| \beta |}  \sqrt{1 + \frac{1}{| \beta |} \| \svec \|^2}, 
    \text{which is satisfied if ~} \| \tvec \|^2 = | \beta |  (1 + \frac{1}{| \beta |} \| \svec \|^2) = | \beta | + \| \svec \|^2.
\end{equation}

Since $\xvec \in \Qcal^{p,q}_\beta$, we have by definition $\| \xvec \|_q^2 = \| \svec \|^2 - \| \tvec \|^2 = \beta < 0$. Therefore, we have:
\begin{align}
    \| \tvec \|^2 = \| \svec \|^2 - \beta = \| \svec \|^2 + | \beta |.
\end{align}

$\bullet$ Let us now show $\psi(\psi^{-1}(\zvec)) = \zvec$. We recall that $\zvec = \begin{pmatrix} \uvec \\ \vvec \end{pmatrix} \in \Scal^{q} \times \RR^{p}$ where $\uvec \in \Scal^{q}$ and $\svec \in \RR^{p}$. 

We need to show:
\begin{equation} \label{eq:u_diffeomorphism}
\uvec =  \frac{\sqrt{| \beta |} \sqrt{1 + \| \vvec \|^2} \uvec}{\| \sqrt{| \beta |} \sqrt{1 + \| \vvec \|^2} \uvec \|} = \frac{\uvec}{ \| \uvec \| }.
\end{equation}
By definition of $\Scal^{q}$, $\uvec$ satisfies Eq. \eqref{eq:u_diffeomorphism}. \qed

\subsection{The metric tensor is positive definite only if $q = 0$ and indefinite in the general case} \label{sec:hyperboloid}

We show here that the hyperboloid (\ie upper sheet of the two-sheet hyperboloid $\hquadricpq{p}{0}{\beta}$ with $p \geq 1$) is a Riemannian manifold even if its symmetric metric matrix $\srbasis = \textbf{I}_{1,p}$ is not positive definite. That result is already known in the literature and a proof can be found in \cite{petersen2006riemannian}. 
We still give it because we think it is important.

We recall that a Riemannian manifold is a pseudo-Riemannian manifold with positive definite metric tensor. 
In other words, for all $\xvec \in \Mcal$ where $\Mcal$ is a pseudo-Riemannian manifold, $\Mcal$ is Riemannian iff $\forall \xii \in T_{\xvec} \Mcal, ~ g_{\xvec} (\xii, \xii) > 0$ iff $\xii \neq 0$. 

We recall that we consider that $\beta < 0$, the usual definition of the hyperboloid $\Hcal^p_{\beta}$ in the literature is:
\begin{equation}
    \Hcal^p_{\beta} =  \left\{ \xvec = \left(x_0, x_1, \cdots, x_{p}\right)^{\top} \in \Qcal^{p,0}_\beta : x_0 \geq 0 \right\},
\end{equation}

Let us note $\xvec = \left(x_0, x_1, \cdots, x_{p}\right)^{\top} \in \Hcal^p_{\beta}$ and $\xii = (\xi_0, \cdots, \xi_p)^{\top}$ a tangent vector at $\xvec$. 
For simplicity, we will denote the vectors $\uvec = \left(x_1, \cdots, x_{p}\right)^{\top} \in \RR^{p}$ and $\vvec = \left(\xi_1, \cdots, \xi_{p}\right)^{\top} \in \RR^{p}$ so that $\xvec = \begin{pmatrix} x_0 \\ \uvec \end{pmatrix} \in \Hcal^p_{\beta}$ and $\xii = \begin{pmatrix} \xi_0 \\ \vvec \end{pmatrix}$. 
By definition of $\Hcal^p_{\beta}$, we have $x_0 = \sqrt{- \beta + \| \uvec \|^2} > 0$. 

Moreover, by definition of the tangent space, and since $\Hcal^p_{\beta}$ is a subset of $\Qcal^{p,0}_\beta$, any tangent vector $\xii$ satisfies $0 = \langle \xvec, \xii \rangle_q = - x_0 \xi_0 + \inner{\uvec}{\vvec}$ where $q = 0$. This implies: 
\begin{equation}
    \xi_0 = \frac{\inner{\uvec}{\vvec}}{x_0} = \frac{\inner{\uvec}{\vvec}}{\sqrt{- \beta + \| \uvec \|^2}} 
    \implies \xi_0^2 = \frac{\inner{\uvec}{\vvec}^2}{- \beta + \| \uvec \|^2}. \label{eq:v_0_value}
\end{equation}
It is obvious that if $\xii = \boldsymbol{0}$, then $\langle \xii, \xii \rangle_q = 0$.  
To show that $\Hcal^p_{\beta}$ is a Riemannian manifold, we need to show that every non-vanishing tangent vector $\xii$ satisfies $\langle \xii, \xii \rangle_q > 0$. 
By definition of $\langle \cdot, \cdot \rangle_q$, we have 
\begin{equation}
    \langle \xii, \xii \rangle_q = - \xi_0^2 + \| \vvec \|^2.
\end{equation}
We then need to show that $\xi_0^2 < \| \vvec \|^2$. 
By the Cauchy-Schwarz inequality, we have $\langle \uvec , \vvec \rangle^2 \leq \| \vvec \|^2 \| \uvec \|^2 \leq \| \vvec \|^2 (- \beta + \| \uvec \|^2)$. This implies from Eq. \eqref{eq:v_0_value}:
\begin{equation}
    \xi_0^2 = \frac{\langle \uvec , \vvec \rangle^2}{- \beta + \| \uvec \|^2} \leq \| \vvec \|^2. \label{eq:cauchy_schwarz}
\end{equation} 
By using the Cauchy-Schwarz inequality, Eq. \eqref{eq:cauchy_schwarz} can be an equality iff $\xii = \boldsymbol{0}$ since $\| \uvec \|^2 < - \beta + \| \uvec \|^2$.  
Since every non-vanishing tangent vector $\xii$ to $\Hcal^p_{\beta}$ satisfies $\langle \xii, \xii \rangle_q > 0$, the metric tensor of $\Hcal^p_{\beta}$ is positive definite. This shows that $\Hcal^p_{\beta}$ is Riemannian. \qed 

The fact that we consider $\beta < 0$ is important. For instance, if $\beta = 1$, then let us note $\xvec = (0,1,0)^{\top} \in \hquadricpq{2}{0}{1}$. 
If $\xii = (1,0,0)^{\top} \in \tangentspace{\xvec} \hquadricpq{2}{0}{1}$, then we have $\langle \xii, \xii \rangle_q < 0$. 
If $\xii = (0,0,1)^{\top} \in \tangentspace{\xvec} \hquadricpq{2}{0}{1}$, then we have $\langle \xii, \xii \rangle_q > 0$. 
Finally, if $\xii = (1,0,1)^{\top} \in \tangentspace{\xvec} \hquadricpq{2}{0}{1}$, then we have $\langle \xii, \xii \rangle_q = 0$. 
The metric tensor is then indefinite if $\beta = 1$, and $\hquadricpq{2}{0}{1}$ is non-Riemannian. 
That result can be extended to any positive value of $\beta$ by considering instead $\xvec = (0, \sqrt{\beta},0)^{\top}$.

\textbf{Indefiniteness:} If $\beta < 0$, one can show in a similar way that $\innerlorentz{\cdot}{\cdot}$ is indefinite if $q \geq 1$ and $p \geq 1$. 
Let us consider $q \geq 1, p \geq 1$ and $\xvec = (0, \sqrt{|\beta|}, \cdots, 0)^{\top} \in \hquadric$ where only the second element of $\xvec$ is nonzero. 
If $\xii \in \tquadric{\xvec}$ is defined such that all its elements except the first one are equal to zero, then $\langle \xii, \xii \rangle_q < 0$. 
If $\xii \in \tquadric{\xvec}$ is defined such that all its elements except the last one are equal to zero, then $\langle \xii, \xii \rangle_q > 0$. The scalar product $\innerlorentz{\cdot}{\cdot}$ is then indefinite if $q \geq 1$ and $p \geq 1$.

\section{Experimental Results} \label{sec:experiments_supp}

We complete here the experimental result section. 
First, we provide additional details about Zachary's karate club dataset experiments in Section \ref{sec:supp_zachary}. 
We provide in Section \ref{sec:nips} similar experimental results on a larger social network dataset about co-authorship at NIPS conferences.

\subsection{Additional details about Zachary's karate club experiments} \label{sec:supp_zachary}

\textbf{Complexity:} We implemented our approach in PyTorch 1.0 \cite{NEURIPS2019_9015} which automatically calculates the Euclidean gradient $\eucgi$ for each node $v_i \in V$. Once $\eucgi$ is computed, the computation of 
$\descent_i = \orthoproj(\srbasis \orthoproj(\srbasis \eucgi))$ is linear in the dimensionality $d$ and is very efficient to compute. The exponential map is also efficient to calculate (\ie linear algorithmic complexity).

\textbf{Running time:} 
The step size/learning rate for all our optimizers is fixed to $\eta = 10^{-6}$ (without momentum). Step size strategies could have been used to improve convergence rate but the goal of our experiment was only to verify that our solvers decrease the optimized function. 

In the weighted case, we stop after 10,000 iterations of descent method. The training process takes 182 seconds ($\sim$3 minutes) on an  Intel i7-7700k CPU, and 254 seconds ($\sim$4 minutes) on an Intel i7-8650U CPU when using the pseudo-Riemannian optimizer introduced in Section \ref{sec:pseudo_riemannian_optimization}. 
The optimizer introduced in Section \ref{sec:diffeomorphisms} is 10\% faster (165 seconds on an Intel i7-7700k CPU) since it requires less computations.

\subsection{Co-authorship from papers published at NIPS} \label{sec:nips}

We also quantitatively evaluate our approach on a dataset that describes co-authorship information from papers published at NIPS from 1988 to 2003 \cite{chechik2007eec}. 

\textbf{Description of the dataset:} The graph $G = (V,E)$ is constructed by considering each author as a node, and an edge $e_k = (v_i, v_j)$ is created iff the authors $i$ and $j$ are co-authors of at least one paper. The capacity $c_k$ is the number of papers co-authored by the pair of authors $e_k = (v_i, v_j)$. 
The original number of authors in the dataset is 2865, but only $n = |V| = 2715$ authors have at least one co-author. The number of edges is $m = |E| = 4733$, which means that the number of pairs of nodes that have no edge joining them is 3,679,522.  The capacity $c_k$ of each edge $e_k$ is a natural number in $\{ 1, \cdots, 9 \}$. 
The ``leaders'' of this dataset are not the authors with highest number of papers, but those with highest number of co-authors.

\textbf{Implementation details and running times:} 
We ran all our experiments for this dataset on a 12 GB NVIDIA TITAN V GPU. 
To fit into memory, when constructing each weaker set $\Wcal(e_k)$, we take into account all the edges with capacity lower than $c_k$ and randomly sample 42,000 pairs of nodes without edge joining them. Our fixed hyperparameters are: $\beta = -1$, the temperature is $\tau = 10^{-5}$, and the step size is $\eta = 10^{-8}$. 
We stop the training of representations lying on $4$-dimensional manifolds (see Table \ref{table:scores_nips}) after 10,000 iterations, which takes 10 hours to train ($\sim1,000$ iterations per hour). 
We stop the training of $6$-dimensional manifolds (see Table \ref{table:scores_nips_6}) after 25,000 iterations because they take longer to converge.

\begin{table}[!t]
  \caption{\label{table:scores_nips} Evaluation scores for the different learned representations lying on 4-dimensional manifolds on the NIPS dataset}
  \centering \scriptsize 
  \begin{tabular}{lcccccc}
    \toprule
    %\multicolumn{2}{c}{Part}                   \\
    %\cmidrule(r){1-2}
    Evaluation metric   & $\RR^4$     & $\Qcal^{4,0}_{-1}$ & $\Qcal^{3,1}_{-1}$ & $\Qcal^{2,2}_{-1}$  & $\Qcal^{1,3}_{-1}$ & $\Qcal^{0,4}_{-1}$ \\ & (flat) & (hyperbolic) & (ours) & (ours) & (ours) & (spherical) \\ \midrule 
    Spearman's $\rho$ for the whole dataset & $0.469$ &  $ 0.460$ &  $0.629$ &  $\textbf{0.667}$ &  $0.625$ & $0.437$ \\
    Spearman's $\rho$ for $s_i \geq 10$ & $0.512$ & $0.490$ & $\textbf{0.552}$ & $0.493$ & $0.441$ & $0.493$ \\ 
        Spearman's $\rho$ for $s_i \geq 20$ & $0.217$ & $0.292$ & $0.316$ & $0.307$ & $0.227$ & $\textbf{0.387}$ \\ 
        Recall@1 (in \%) & $70.7$ & $70.2$ & $\textbf{83.5}$ & $82.6$ & $78.1$ & $69.7$\\ 
    \bottomrule
  \end{tabular}

  \caption{\label{table:scores_nips_6} Evaluation scores for the different learned representations lying on 6-dimensional manifolds on the NIPS dataset}
  \centering \scriptsize 
  \begin{tabular}{lcccccccc}
    \toprule
    %\multicolumn{2}{c}{Part}                   \\
    %\cmidrule(r){1-2}
    Evaluation metric   & $\RR^6$     & $\Qcal^{6,0}_{-1}$ & $\Qcal^{5,1}_{-1}$ & $\Qcal^{4,2}_{-1}$  & $\Qcal^{3,3}_{-1}$ & $\Qcal^{2,4}_{-1}$ & $\Qcal^{1,5}_{-1}$ & $\Qcal^{0,6}_{-1}$ \\ & (flat) & (hyperbolic) & (ours) & (ours) & (ours) & (ours) & (ours) & (spherical) \\ \midrule 
    Spearman's $\rho$ for the whole dataset & $0.554$ &  $ 0.455$ &  $0.617$ &  $0.666$ &  $\textbf{0.688}$ & $0.675$ & $0.610$ & $0.456$ \\
    Spearman's $\rho$ for $s_i \geq 10$ & $0.459$ & $0.515$ & $0.507$ & $0.501$ & $0.514$ & $0.470$ & $0.494$ & $\textbf{0.532}$ \\ 
        Spearman's $\rho$ for $s_i \geq 20$ & $0.372$ & $0.321$ & $0.211$ & $0.317$ & $\textbf{0.383}$ & $0.370$ & $0.213$ & $0.257$ \\ 
        Recall@1 (in \%) & $88.6$ & $\textbf{99.9}$ & $93.1$ & $96.4$ & $96.7$ & $95.7$ & $93.5$ & $\textbf{99.9}$\\ 
    \bottomrule
  \end{tabular}  
\end{table}

\textbf{Quantitative evaluation:} The evaluation process is the same as the evaluation for Zachary's karate club dataset. 
However, unlike for Zachary's karate club dataset, the factions and their leaders are unknown. We then only consider Spearman's rank correlation coefficient scores. We also consider the average recall at 1 as explained below.

The capacity matrix $\Capacity \in \{0,1,\cdots,8,9\}^{n \times n}$ is symmetric, so our score matrix is $\SymCapacity = \Capacity$. We select the most influential members based on the score $s_i = \sum_{j=1}^n \SymCapacity_{ij} \in \{ 1, \cdots, 97 \}$. %This score takes one of the following values: 
%1, 2,     3,     4,     5,     6,     7,     8,     9,    10,    11,    12,    13,    14,
% 15,    16,    17,    18,    19,    20,    21,    22,    23,    24,    25,    26,    27,    28,
%30,    31,    32,    34,    35,    37,    38,    39,    40,    46,    48,    49,    64,    70, 
%77,    97. 
Spearman's rank correlation coefficient \cite{spearman2015proof} between the selected ordered scores $s_i$ and corresponding $\delta_i  = \sum_{j=1}^n \dsf(\xvec_i, \xvec_j)$ is reported in Table \ref{table:scores_nips} and shows the relevance of our representations. 
The most influential members are selected if their score $s_i$ is at least $1$ (\ie top $n$ members where $n = 2715$ is the number of nodes), at least $10$ (\ie top 232 authors) and at least $20$ (\ie top 57 authors). 
We also report the average recall at 1 (in percentage): for each node $v_i$, we find the nearest neighbor $v_j$ \wrt the chosen metric, the recall at 1 is $0$ if $\Capacity_{ij} = 0$, and $1$ otherwise.  
Both $Q_{-1}^{3,1}$ $Q_{-1}^{2,2}$ outperform the hyperbolic model.

If we use the Euclidean norm of hyperbolic representations (in $\Qcal^{4,0}_{-1}$) as proxy, we obtain the following scores: 
0.044, -0.113, 0.264 for the top 2715, to 232 and top 57 members respectively. 
These scores are worse than using $\delta_i$ as proxy due to the presence of cycles in the graph.

In the 6-dimensional manifold case, the Recall@1 score is better for the hyperbolic and spherical representations. On the other hand, the ultrahyperbolic representations perform better in terms of Spearman's rank correlation coefficient.

In conclusion, our proposed ultrahyperbolic representations extract hierarchy better than hyperbolic representations when the hierarchy graph contains cycles.

\end{document}